\documentclass[]{article}   
\usepackage{proceed2e} 

\usepackage{fontenc}
\usepackage[margin=2cm, a4paper]{geometry}

\usepackage{amsmath,amsthm,amsfonts,amssymb}
\usepackage[all]{xy}
\usepackage{graphicx,subfigure,multirow}
\usepackage[figurename=Fig.]{caption}
\usepackage{tikz}
\usetikzlibrary{shapes, arrows, shadows}
\usepackage{pdflscape}
\usepackage{tabularx}
\usepackage{comment}
\usepackage{hyperref}

\newtheorem{thm}{Theorem}[subsection]
\newtheorem{define}[thm]{Definition}

\newtheorem{lem}[thm]{Lemma}

\numberwithin{equation}{subsection}

\usepackage{times} 

\title{Causal inference via algebraic geometry: feasibility tests for functional causal structures with two binary observed variables}

\author{ {\bf{Ciar{\'a}n~M.~Lee}\thanks{Electronic address: ciaran.lee@cs.ox.ac.uk. }} \\
Department of Computer Science,\\
University of Oxford, \\
Oxford, UK OX1 3QD \\ 
\And
{\bf {Robert~W.~Spekkens}\thanks{Electronic address: rspekkens@perimeterinstitute.ca. }}  \\
Perimeter Institute for Theoretical Physics,        \\ 
Waterloo, Ontario,\\
Canada N2L 2Y5
}

\begin{document}

\maketitle

\begin{abstract} 
We provide a scheme for inferring causal relations from uncontrolled statistical data based on tools from computational algebraic geometry, in particular, the computation of Groebner bases. We focus on causal structures containing just two observed variables, each of which is binary.  
We consider the consequences of imposing different restrictions on the number and cardinality of latent variables and of assuming different functional dependences of the observed variables on the latent ones (in particular, the noise need not be additive). 
We provide an inductive scheme for classifying functional causal structures into distinct observational equivalence classes. For each observational equivalence class, we provide a procedure for deriving constraints on the joint distribution that are necessary and sufficient conditions for it to arise from a model in that class. We also demonstrate how this sort of approach provides a means of determining which causal parameters are identifiable and how to solve for these. Prospects for expanding the scope of our scheme, in particular to the problem of quantum causal inference, are also discussed.  
\end{abstract} 


\section{Introduction}








Causal relationships, unlike statistical dependences, support inferences about the effects of interventions and the truths of counterfactuals.  While a randomised controlled experiment can be used to determine causal relationships, these may not be available for various reasons: they could be restrictively expensive, technologically infeasible, unethical (e.g., assessing the effect of smoking on lung cancer), or indeed physically impossible (e.g., for variables describing properties of distant astronomical bodies). Therefore, inferring causal relationships from uncontrolled statistical data is an important problem, with broad applicability across scientific disciplines. Over the past-twenty five years, there has been much progress in developing methods to solve this problem \cite{P, Sprite, Confounders, Noise, Discrete}. 

As has become standard practice, we formalize the notion of causal structure using directed acyclic graphs (DAGs) with random variables as nodes and arrows representing direct causal influence~\cite{P, Sprite}.  \color{black}  A more refined description of causal dependences specifies not only what causes what, but also, for every variable, its functional dependence on its causal parents.  We shall use the term {\em functional causal structure} to refer to the specification of the set of functions, which includes a specification of the DAG.  
As is standard, the variables that are not observed are termed {\em latent}, and the DAG does not include any latent variables that act as causal mediaries, so that all the latent variables are parentless.  We shall use the term {\em causal model} to describe the functional causal structure together with a specification, for each latent variable, of a probability distribution over its values. 
Each causal model associated to a given functional causal structure defines a possible joint probability distribution over the observed variables.  We are interested in the set of possible joint distributions over the observed variables for a given functional causal structure, that is, those that can arise from {\em some} set of distributions on the latent variables.
\color{black} 
We will say that two functional causal structures are {\em observationally equivalent} if they are characterized by the same set of distributions over the observed variables.\footnote{This should not be confused with the notion of observational equivalence as applied to DAGs~\cite{P}.} 
 
Many causal inference algorithms, such as those of \cite{P} and \cite{Sprite}, only make use of conditional independence relations among the observed variables. 
\color{black} If two causal structures are such that the same set of conditional independence relations are faithful to them, then they are said to be Markov equivalent. \color{black}
\color{black}
Note that Markov equivalence can be decided purely on the basis of the DAG (i.e., the causal structure), while the notion of observational equivalence of interest here depends on the functional dependences (i.e., the {\em functional causal structure}).  
In the case of just two observed variables, which is the one we consider here, the set of all causal structures are partitioned into just two Markov equivalence classes: those wherein the variables are causally connected, and those wherein they are not. 
As we show, however, the joint distribution over the observed variables supports many more inferences about the functional causal structure, thereby providing a more fine-grained classification than is provided by Markov equivalence.  \color{black}



In recent years, several methods have been suggested that make use not only of conditional independences, but also other properties of the joint statistical distribution between the observed variables \cite{Confounders, Noise, Discrete, CANM} (See also the works discussed in Secs.~\ref{quantum} and \ref{related}). These newer methods also have limitations in the sense that they impose restrictions on the number of latent variables allowed in the underlying causal model and also on the mechanisms by which these latent variables influence the observed ones. 

In the present work, we restrict attention to the causal inference problem where there are just two observed variables, each of which is binary (that is, discrete with just two possible values).  We allow any functional causal structure involving 
latent variables that are discrete (with a finite number of values), and we impose no restriction on the number of latent variables or the mechanisms by which these influence the observed ones.  



We provide an inductive scheme for characterizing the observational equivalence classes of functional causal structures. This scheme has a few steps.  First we show that, in each observational class, there is a functional causal structure wherein all of the latent variables are binary.  
Restricting ourselves to the latter sort of functional causal structure, 
we show that one can inductively build up any functional causal structure from a pair of others having fewer latent variables. Thus, starting with functional causal structures with no latent variables, we can recursively build up all functional causal structures, and therefore all observational equivalence classes of these, by applying our inductive scheme.

Using this scheme, we catalogue all observational equivalence classes generated by functional causal structures with four or fewer binary latent variables. We have evidence, but no proof yet, that our catalogue is complete in the sense that a functional causal structure with any number of binary latent variables---and hence, by the connection described above, any functional causal structure with discrete latent variables---belongs to one of the classes we have identified. 

We also describe a procedure for deriving, for each class, the set of necessary and sufficient conditions on the joint distribution over observed variables for it to be possible to generate it from functional causal structures in this class.  We call such a set of conditions a {\em feasibility test} for the class.   
The procedure for deriving these is as follows.
We start with a particular functional causal structure within the class, express the parameters in the joint probability distribution over the observed variables 
in terms of the parameters in the probability distributions over 
the latent variables, then eliminate the latter using techniques from algebraic geometry.   
 
Finally, we consider applications to the problem of identifying causal parameters.  For the parameters describing the probability distributions over the latent variables, we note that our technique allows one to find expressions for these in terms of the observational data for each observational equivalence classes that we have considered.    For the parameters describing the functional relations, we note that the limits to what one can infer about these, which may be different for different points in the space of possible joint distributions over the observed variables, can be inferred from our feasibility tests.
 \color{black}

\section{Setting up the problem} \label{one} 

\begin{figure}[t]
\begin{subfigure}
\centering
(a)\includegraphics[scale=0.59]{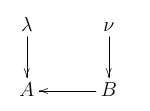}
\end{subfigure}
\quad
\begin{subfigure}
\centering
(b)\includegraphics[scale=0.3]{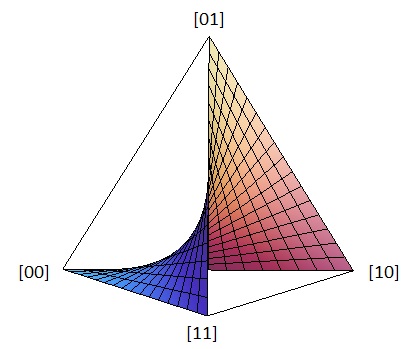}
\end{subfigure}
\caption{
(a) DAG for causal model defined by $A=B\oplus\lambda$ and $B=\nu$ (b) Joint distributions that can be generated by this causal model. 
}
\label{ANM1}
\end{figure}

Consider the causal model of Fig.\ref{ANM1}(a).  From the DAG, it is clear that $B$ is a cause of $A$, while $\lambda$ is noise local to $A$ and $\nu$ is noise local to $B$. 
The functional dependences are given by $A=B\oplus\lambda$ and $B=\nu$.  
A model with this sort of functional dependence is referred to as an \emph{additive noise model} (ANM) in Refs.~\cite{Confounders, Noise, Discrete,  CANM}. The values of $A$, for different values of $B$ and $\lambda$, are given in the table below.
\begin{center}
    \begin{tabular}{ | l | l | l | l |} 
    \hline
    $\nu$ & $\lambda$ & $B$ & $A=B\oplus\lambda$ \\ \hline
    $0$ & $0$ &  $0$& $0$ \\ \hline
    $0$ & $1$ & $0$ & $1$  \\ \hline
    $1$ & $0$ & $1$ & $1$  \\ \hline
    $1$ & $1$ & $1$ & $0$  \\ \hline
        \end{tabular}
\end{center} 

\begin{figure}[t]
\begin{centering}
\begin{subfigure}
\centering
(a) \includegraphics[scale=0.59]{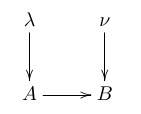}
\end{subfigure}
\qquad
\begin{subfigure}
\centering
(b)\includegraphics[scale=0.22]{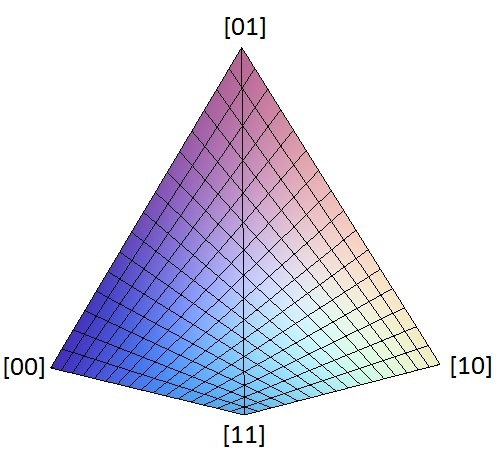}
\end{subfigure} \quad  
\begin{subfigure}
\centering
\qquad (c)\includegraphics[scale=0.59]{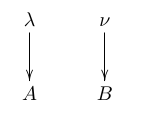} 
\end{subfigure}
\qquad
\begin{subfigure}
\centering
(d)\includegraphics[scale=0.31]{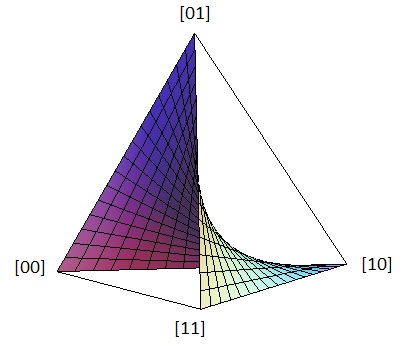}
\end{subfigure}
\caption{(a) DAG for causal model defined by $A=\lambda$ and $B=A\oplus\nu$. (b) Joint distributions that can be generated by the model of (a). \color{black} Note that this is a head-on view of a fan shape of the same type as is depicted in (d). \color{black}
(c) DAG for causal model defined by $A=\lambda$ and $B=\nu$. (d) Joint distributions that can be generated by the model of (c).}
\label{ANM2}
\end{centering}
\end{figure}

In Ref.~\cite{Discrete}, it was shown that one can distinguish between the causal model of Fig.\ref{ANM1}(a) and the causal models depicted in Fig.~\ref{ANM2}(a) and Fig.~\ref{ANM2}(c), except for special cases of the distributions over the noise variables, such as, for instance, when $\lambda$ and $\nu$ are uniformly distributed.
Thus if we are promised that the causal model is an ANM, 
then (except for the special cases) we can distinguish between $B$ causing $A$, $A$ causing $B$ and $A$ and $B$ being causally disconnected. To see how this works we will need to determine the correlations generated by this model. 

To describe the correlations we adopt the following notational convention. $$\begin{aligned} 
\mathbb{P}(A)&=[x] \quad\mathrm{means}\quad \mathbb{P}(A=x)=1\\
\mathbb{P}(A,B)&=[x][y]=[xy] \quad\mathrm{means}\quad \mathbb{P}(A=x,B=y)=1\\
\mathbb{P}(A)&=q[x] \quad \mathrm{means}\quad \mathbb{P}(A=x)=q.
\end{aligned}$$

Let $q_1$ be the probability that $\nu=0$ and $q_2$ be the probability that $\lambda=0$, then the correlations for the above causal model are $$\begin{aligned} \mathbb{P}(A,B)=q&_1q_2[00]+(1-q_1)(1-q_2)[01] \\ &+q_1(1-q_2)[10]+(1-q_1)q_2[11], \end{aligned}$$ This means $\mathbb{P}(A=0,B=0)=q_1q_2,$ $\mathbb{P}(A=0,B=1)=(1-q_1)(1-q_2)$ and so on. From now on, we will use the shorthand 
$\overline{q}_i \equiv 1-q_i$ to simplify expressions.

\color{black} Note that if a latent variable were to take one of its values with probability 1, then it would be trivial and could be eliminated from the functional causal structure.  We therefore consider only functional causal structures with nontrivial latent variables, that is, latent variables that have some statistical variation in their value, so that the probability of any value is bounded away from 0 and 1.  In the present example, therefore, $0< q_1, q_2<1$. \color{black}

For a general causal model we have $\mathbb{P}(A,B)=p_{00}[00]+p_{01}[01]+p_{10}[10]+p_{11}[11]$, where $\mathbb{P}(A=i,B=j)=p_{ij}$. We note that $p_{00}+p_{01}+p_{10}+p_{11}=1$. As we only need three real parameters to specify $\mathbb{P}(A,B)$, we can plot it in $\mathbb{R}^3$. It is easy to see that the points $\{\mathbb{P}(A=i,B=j)=1:i,j\in\mathbb{Z}_2\}$ form the vertices of a tetrahedron in $\mathbb{R}^3$ and so the plot of $\mathbb{P}(A,B)$ must lie within 
this tetrahedron.

We can rewrite $\mathbb{P}(A,B)$ for our current example as
$$\mathbb{P}(A,B)=q_1\big(q_2[00]+\overline{q}_2[10]\big)+\overline{q}_1\big(q_2[11]+\overline{q}_2[01]\big).$$
\color{black} So, if we fix the value of $q_2$ in the range $(0,1)$ and vary $q_1$ over the interval $(0,1)$, the plot of $\mathbb{P}(A,B)$ consists of the line passing through a point on the edge of the tetrahedron containing the vertices $\{[00],[10]\}$ and a point on the edge containing the vertices $\{[11],[01]\}$ (but excluding these points).  The full plot of $\mathbb{P}(A,B)$, as  $q_1$ and $q_2$ each range  over the interval $(0,1)$, is depicted in Fig.~\ref{ANM1}(b) (where  the boundary points are excluded). We refer to this shape as a \emph{fan}. Fig.~\ref{ANM2}(b) and Fig.~\ref{ANM2}(d) depict the set of joint distributions for the ANM where $A$ causes $B$ and the causal structure where $A$ and $B$ are causally disconnected. \color{black}

Given some joint distribution, $\mathbb{P}(A,B)$, how do we determine if it lies on one of the fans of Fig.~\ref{ANM1}(b), Fig.~\ref{ANM2}(b) or Fig.~\ref{ANM2}(d)? Recall that, because the latent variables are unobserved, we do not have access to the $q_i$'s directly, only the observed $p_{ij}$'s. Thus, the problem can be posed as follows: what are the defining equations of the fans in terms of the observed $p_{ij}$'s?

This problem was solved for the example of Fig.~\ref{ANM1} in Ref.~\cite{Discrete} using the following technique.  First, it was noted that the DAG implies that $\lambda$ is marginally independent of $B$, and therefore $\mathbb{P}(\lambda|B=0)=\mathbb{P}(\lambda|B=1)$.  Given that $\lambda$ is a binary variable, this is true if and only if $\mathbb{P}(\lambda=1|B=0)=\mathbb{P}(\lambda=1|B=1)$.  We wish to eliminate $\lambda$ from this condition.  Recall from the definition of conditional probability that  $\mathbb{P}(\lambda=1|B=b) =\mathbb{P}(\lambda=1,B=b)/\mathbb{P}(B=b)$. The functional dependence $A = B \oplus \lambda$ can be used to 
conclude that $\mathbb{P}(\lambda=1,B=b)=\mathbb{P}(A=b\oplus 1,B=b)$.  Note that this last step is only possible because the noise is additive, so that one can infer $\lambda$ from $A$ and $B$. Therefore, reverting to our notational conventions, where $\mathbb{P}(A=1\oplus b,B=b) = p_{1\oplus b, b}$ and $\mathbb{P}(B=b)= p_{0,b} + p_{1,b}$, the condition becomes
$$   
\frac{p_{10}}{p_{00}+p_{10}}=\frac{p_{01}}{p_{11}+p_{01}},$$ which can be rewritten as:
$${p_{00}p_{01}=p_{11}p_{10}}.
$$
\color{black} This equation, together with the {\em open-interval constraints}, $$0<p_{00},p_{01},p_{10},p_{11}<1,$$
defines the fan in Fig.~\ref{ANM1}(b).  \color{black}
Using similar techniques, one can show that Figs.~2(b) and ~2(d)  are defined by equation
$${p_{00}p_{10}=p_{11}p_{01}},$$ respectively
$${p_{00}p_{11}=p_{10}p_{01}},$$ together with the open-interval constraint. \color{black}





The question is: how can one find feasibility tests for generic causal models?  In particular, how does one treat models where the noise is not additive?
Consider, for instance, the causal model that has the same DAG as in Fig.~\ref{ANM1}(a), but where the noise is multiplicative, that is, $A=B\lambda$.  In this case, the value of $\lambda$ cannot be inferred from $A$ and $B$ (given that these could be zero), and consequently one cannot use the approach of Ref.~\cite{Discrete}.
It is also unclear how one can characterize the possibilities for the joint distribution when the causal model involves an arbitrary number of latent variables.
We will show that these questions can be answered 
using powerful tools from algebraic geometry, which we describe in the next section. 

\section{Deriving the feasibility tests} \label{two}  

We begin with an introduction to some of the main concepts of algebraic geometry following the presentation 
given in \cite{geo}. For a more detailed discussion, see appendix~\ref{alg geo}.  


Denote the set of all polynomials in variables $x_1,\dots,x_n$ with coefficients in some field $k$ by $k[x_1,\dots,x_n]$.
When dealing with polynomials, we are mainly interested in the solution set of systems of polynomial equations. This leads us to the main geometrical objects studied in algebraic geometry, algebraic varieties and semi-algebraic sets.


An {\em algebraic variety}\footnote{Also called an {\em affine variety} \color{black} or an {\em algebraic set}.\color{black}} $\bold{V}(f_1,\dots,f_s)\subset{k^n}$ is the solution set of the system of polynomial equations $f_1(x_1,\dots,x_n)=\dots=f_s(x_1,\dots,x_n)=0.$  
\color{black} A \emph{basic semi-algebraic set} is defined to be the solution set of a system of polynomial equalities and inequalities, that is, $\{x\in \mathbb{R}^n: g_i(x)\rightleftharpoons 0, \text{ }\forall \text{ } i=1,\dots, m\},$ where$g_1,\dots,g_n \in \mathbb{R}[x_1,\dots,x_n]$ are polynomials ove the reals\footnote{Note that one can replace the real field $\mathbb{R}$ used in the last definition with any \emph{ordered} field.} and where $\rightleftharpoons \text{corresponds to either } \geq,\text{ } =, \text{ or } \leq$.  Note that algebraic varieties are examples of basic semi-algebraic sets.  A \emph{semi-algebraic set} is formed by taking finite combinations of unions, intersections, or complements of basic semi-algebraic sets. 
For instance, the fan in Fig.1(b) is the semi-algebraic set that results from the intersection of the algebraic variety defined by the single polynomial equation $p_{00}p_{01}-p_{11}p_{10} =0$ and the set of inequalities that define the interior of the tetrahedral probability simplex (requiring each probability to be in the interval $(0,1)$). \color{black}

More generally, for {\em any} causal model, the set of possible joint distributions that can be generated by it are represented by a semi-algebraic set.   It follows that two causal models are observationally equivalent if and only if they generate the same semi-algebraic set.

We now define \emph{ideals}, the main algebraic object studied in algebraic geometry. 
A subset $I\subset{k[x_1,\dots,x_n]}$ is an \emph{ideal} if it satisfies: (1) $0\in{I}$, (2) If $f,g\in{I}$, then $f+g\in{I}$,
and (3) If $f\in{I}$ and $h\in{k[x_1,\dots,x_n]}$, then $hf\in{I}$.

A natural example of an ideal is the ideal generated by a finite number of polynomials, defined as follows. Let $f_1,\dots,f_s$ be polynomials in ${k[x_1,\dots,x_n]}$, then the \emph{ideal generated by} $f_1,\dots,f_s$ is:
$$\langle{f_1,\dots,f_s}\rangle=\Big\lbrace\sum_{i=1}^sh_if_i:h_1,\dots,h_s\in{k[x_1,\dots,x_n]}\Big\rbrace.$$ The polynomials $f_1,\dots,f_s$ are called the \emph{basis} of the ideal.


Studying the relations between certain ideals and varieties forms one of the main areas of study in algebraic geometry. One can even define the algebraic variety $\bold{V}(I)$ defined by the ideal $I\subset{k[x_1,\dots,x_n]}$, where 
$$\bold{V}(I)=\{(a_1,\dots,a_n)\in{k^n}:f(a_1,\dots,a_n)=0,\forall{f\in{I}}\}.$$ Interestingly, it can also be shown that if $I=\langle{f_1,\dots,f_s}\rangle$, then $\bold{V}(I)=\bold{V}(f_1,\dots,f_s)$, \color{black}  which is to say that the variety defined by a set of polynomials is the same as the variety defined by the ideal generated by those polynomials. Hence, \emph{varieties are determined by ideals}. \color{black}

We can now use the language of algebraic geometry to restate the question asked at the end of the last section. Let $V\subseteq k^n$ be an algebraic variety given parametrically as
\begin{equation} \label{poly}
\begin{aligned}
p_1&=g_1(q_1,\dots,q_m),\\
&\vdots\\
p_n&=g_n(q_1,\dots,q_m),
\end{aligned} 
\end{equation}
where the $g_i$ are polynomials in $q_1,\dots,q_m$. \color{black} The conjunction of the above equalities with the inequalities ensuring that the variables $q_1, \dots, q_m$ are  in the interval $(0,1)$ (probabilities bounded away from 0 and 1) defines a semi-algebraic set on $p_1,\dots,p_n, q_1, \dots, q_m$. 
 \color{black} We seek to infer which values of $p_1,\dots,p_n$ are possible for {\em some} values of the $q_1, \dots, q_m$ in their allowed intervals.  By the Tarski-Seidenberg theorem~\cite{TarskiSeidenberg}, the solution to this problem is also a semi-algebraic set.  We determine the latter as follows.  First, we eliminate the variables $q_1, \dots, q_m$ to find a system of polynomial equations in $p_1,\dots,p_n$,.  These define the smallest algebraic variety on $p_1,\dots,p_n, q_1, \dots, q_m$ that contains the semi-algebraic set that we seek to characterize.  This problem is known as \emph{implicitization}.  The second step is to determine which points in this algebraic variety can be extended to a solution of the equalities and inequalities of the original parametric characterization.

\color{black}


For example, consider the algebraic variety that is defined parametrically by the polynomial equations $$p_{00}=q_1q_2,\quad p_{10}=q_1\overline{q}_2,\quad p_{01}=\overline{q}_1\overline{q}_2,\quad p_{11}=\overline{q}_1q_2.$$  We would like to characterize the semi-algebraic set that this variety defines on the observed variables $p_{00},p_{01},p_{10},p_{11}$ alone when one eliminates the parameters $q_1$ and $q_2$ while enforcing that they are probabilities in $(0,1)$.  In Sec.~\ref{one}, it was shown how one can do so, and that the resulting semi-algebraic set is the one depicted in Fig.1(b).
However, the technique was not generalizable to arbitrary functional causal structures.  Here, we reconsider this example using techniques that are generally applicable.

The problem can be solved by employing a specific choice of basis for the ideal generated by the system of polynomial equations that define the variety (\ref{poly}). The basis that achieves this feat is known as the \emph{Groebner basis}.

 

Groebner bases simplify many calculations in algebraic geometry and they have many interesting properties \cite{geo}. There are efficient algorithms for calculating Groebner bases and many software packages that one can use to implement them.

 
We discovered in this section that the fan of Fig.~\ref{ANM1}(b) \color{black} is in fact the intersection of the algebraic variety defined by the ideal $$\langle{p_{00}-q_1q_2, p_{10}-q_1\overline{q}_2, p_{01}-\overline{q}_1\overline{q}_2, p_{11}-\overline{q}_1q_2}\rangle$$ with the tetrahedron.\color{black}  The Groebner basis \footnote{with respect to the lex order $q_1>q_2>p_{00}>p_{10}>p_{01}>p_{11}$, see appendix~\ref{alg geo}} of this ideal is found to be
$$\begin{aligned}
g_1&=q_1+p_{01}+p_{11}-1\\
g_2&=q_2+p_{01}+p_{10}-1\\ 
g_3&=p_{00}+p_{01}+p_{10}+p_{11}-1\\
g_4&=p_{01}^2+p_{01}p_{10}+p_{11}p_{01}-p_{01}+p_{10}p_{11}.
\end{aligned}$$
Solutions to $g_1=\dots=g_4=0$ provide solutions to $$p_{00}=q_1q_2,\quad p_{10}=q_1\overline{q}_2,\quad p_{01}=\overline{q}_1\overline{q}_2, \quad p_{11}=\overline{q}_1q_2$$ which define our algebraic variety. Looking more closely at the Groebner basis we note that the variables $q_1,q_2$ have been eliminated from the polynomials $g_3$ and $g_4$. The solution of $g_3=p_{00}+p_{01}+p_{10}+p_{11}-1=0$ is exactly the normalisation condition. The solution of $g_4=0$ gives us the following $$p_{01}\big(p_{10}+p_{01}+p_{11}-1\big)+p_{10}p_{11}=0,$$ which, using the normalization condition, then gives us
$$p_{00}p_{01}=p_{10}p_{11}.$$
\color{black} On demanding $0<p_{00}, p_{01}, p_{10}, p_{11}<1$ and $p_{ij}\in\mathbb{R}, \forall {ij}$ (i.e. on taking the intersection of this algebraic variety with the tetrahedron), \color{black} we obtain the semi-algebraic set corresponding to the fan of Fig.1(b), which we derived in section \ref{one}.  \color{black}  
This is a special case of a general result, known as the \emph{elimination theorem}, which provides us with a way of using Groebner bases to systematically eliminate certain variables from a system of polynomial equations and, thus, to solve the implicitization problem. 


\color{black} The general procedure for finding the semi-algebraic set is as follows. \color{black}
First, given the system of polynomial equations defining the implicitization problem,
as in Eq.~(\ref{poly}), form the ideal generated by these polynomials and compute\footnote{with respect to the lexicographic order $q_1>q_2>\dots>q_m>p_1>\dots>p_n.$} its Groebner basis.
 \color{black} The elements of this basis that do not contain the variables $q_1,\dots,q_m$ constitute constraints on the variables $p_1, \dots, p_n$ alone. These constraints consitute polynomial equalities, and therefore define an algebraic variety on the variables $p_1, \dots, p_n$. Second, we determine which points on this variety correspond to solutions of the original equalities and inequalities on $q_1,\dots, q_n$ and $p_1, \dots, p_m$.  This will result in inequality constraints. 
 The equality constraints from the first step and the inequality constraints from the second step together characterize the semi-algebraic set on $p_1,\dots, p_n$ that is compatible with the given functional causal structure.   We note that one trivial consequence of the fact that each of the $q_1, \dots, q_m$ is in the interval $(0,1)$ is that each of the $p_1,\dots, p_n$ is in the interval $(0,1)$.  As such, the semi-algebraic set we seek to characterize is always a subset of the geometric intersection of the algebraic variety we find in the first step and the probability simplex on $p_1,\dots, p_n$.  Note, however, that it is generically a strict subset of this intersection. 
  \color{black}

These \color{black} inequality \color{black} constraints manifest themselves in different ways. We present an example of one such manifestation below and leave the remaining examples to appendix~\ref{examples}.



\begin{figure}[t]
\begin{centering} 
\begin{subfigure}
\centering
(a)\includegraphics[scale=0.42]{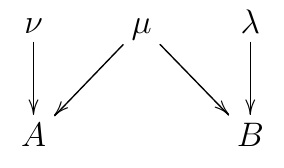}  
\end{subfigure}
\qquad
\begin{subfigure}
\centering
(b)\includegraphics[scale=0.34]{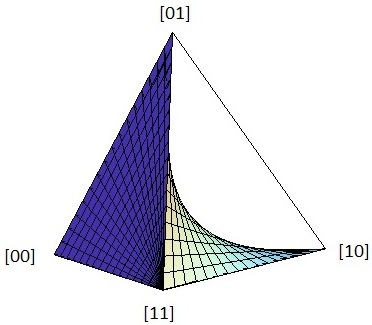}    
\end{subfigure}
\caption{(a) $A=\mu\nu$ and $B=\mu\lambda$. (b) $p_{00}p_{11}\geq{p_{01}p_{10}}$.}
\label{filledfan}
\end{centering}
\end{figure}

Consider the causal model of Fig.~\ref{filledfan}(a).  Defining $q_1$, $q_2$ and $q_3$ to be the probabilities for $\mu=0$, $\nu=0$ and $\lambda =0$ respectively, the joint distribution generated by this model is 
\begin{align} \mathbb{P}(A,B)=(q_1&+\overline{q}_1q_2q_3)[00]+\overline{q}_1q_2\overline{q}_3[01]\nonumber\\
&+\overline{q}_1\overline{q}_2q_3[10]+\overline{q}_1\overline{q}_2\overline{q}_3[11].
\label{mmm}\end{align} 


 \color{black} We begin by providing an intuitive account of the semi-algebraic set describing such joint distributions.  Note first that $\mathbb{P}(A,B)$ can be rewritten as 
$$\begin{aligned} \mathbb{P}(A,B)=q_1[00]+\overline{q}_1\Big(q_2q_3&[00]+q_2\overline{q}_3[01]\\&+\overline{q}_2q_3[10]+\overline{q}_2\overline{q}_3[11]\Big),
\end{aligned}$$ 
implying that it is the convex combination, with weight $q_1$, of the point distribution $[00]$, and with weight $\overline{q}_1$, of the distribution arising from the functional causal structure of Fig.~\ref{ANM2}(c), shown above to be characterized by the equality $p_{00}p_{11}=p_{10}p_{01}$.
It follows that the semi-algebraic set defined by $\mathbb{P}(A,B)$ contains all interior points on any line extending from the vertex $[00]$ to a point on the fan depicted in Fig.~\ref{ANM2}(d); this variety is  depicted in Fig.~\ref{filledfan}(b). 
\color{black} 

\color{black} Reading off the expressions for $p_{00}, p_{01}, p_{10}$, and $p_{11}$ from Eq.~\eqref{mmm}, we obtain the set of polynomials that define the full algebraic variety. The ideal generated by these is:\color{black}
$$\begin{aligned}\langle p_{00}-q_1-\overline{q}_1q_2q_3,p_{01}-\overline{q}_1q_2\overline{q}_3,p_{10}&-\overline{q}_1\overline{q}_2q_3, \\ &p_{11}-\overline{q}_1\overline{q}_2\overline{q}_3 \rangle.\end{aligned}$$
\color{black} To implement the first step of the general procedure outlined above, we derive the Groebner basis for this ideal \footnote{with respect to the lex order $q_1>q_2>q_3>p_{00}>p_{01}>p_{10}>p_{11}$}:
$$\begin{aligned}
g_1&=q_2q_1-q_1-q_2-p_{10}-p_{11}+1\\
g_2&=q_3q_1-q_1-q_3-p_{01}-p_{11}+1 \\
g_3&=q_3p_{10}+q_3p_{11}-p_{10}\\
g_4&=q_2p_{01}+q_3p_{11}-p_{01}\\
g_5&=p_{00}+p_{01}+p_{10}+p_{11}-1\\
g_6&=p_{11}^2+p_{01}p_{10}+p_{11}p_{10}-p_{11}+p_{01}p_{11}+p_{11}q_1.
\end{aligned}$$

Now $g_5=0$ is just the normalisation condition and $g_6=0$ gives the following: 
$$p_{11}\big(p_{10}+p_{01}+p_{11}-1\big)+p_{01}p_{10}+p_{11}q_1=0$$
which, using the normalisation condition, results in
\color{black}
\begin{align}\label{identification1}
q_1 =\frac{p_{11}p_{00}-p_{10}p_{01}}{p_{11}}.
\end{align}
\color{black}

To implement the second step of our procedure, we begin by enforcing $q_1> 0$. This results in the following inequality $$p_{11}p_{00}>{p_{10}p_{01}}.$$ None of the remaining constraints $0 < q_i < 1$, for $i\in \{2,3\}$ result in nontrivial relations among the $p_{ij}$'s, \color{black} \color{black} so the latter inequality is the {\em only} nontrivial constraint.  Together with the open-interval constraints $0<p_{00},p_{01},p_{10},p_{11}<1,$ it describes the necessary and sufficient conditions for the distribution on observed variables to be compatible with the functional causal structure of Fig.~\ref{filledfan}(a).  These conditions define the semi-algebraic set depicted in Fig.~\ref{filledfan}(b).  \color{black}





\section{Characterizing the observational equivalence classes}   \label{characterizing}


In this section, we will provide a scheme for inductively characterizing all observational equivalence classes. As noted in the introduction, we consider only causal models where there is a pair of binary observed variables, which we denote by $A$ and $B$. 


\subsection{Sufficiency of considering purely common-cause models}

A causal model having no directed causal influences between the observed variables will be termed {\em purely common-cause}.   
\begin{lem} 
Every causal model wherein there is a directed causal influence between $A$ and $B$ (either $A \to B$ or $B \to A$) is observationally equivalent to one that is purely common-cause. 
\end{lem}
The proof  is as follows.  Suppose that there is a directed causal influence $B \to A$.  If the collection of all latent variables is denoted by $\lambda$, then a general causal model can be specified by the functional dependences $B = f(\lambda)$ and $A=g(\lambda,B)$ for some functions $f$ and $g$.  But this is observationally equivalent to the causal model that is purely common-cause with functional dependences $B = f(\lambda)$ and $A=g'(\lambda)$ where $g'(\lambda)\equiv g(\lambda,f(\lambda))$. 
In characterizing the distinct observational equivalence classes, therefore, it suffices for us to consider the models that are purely common-cause, and therefore we restrict  our attention to these henceforth.

\begin{figure}[t]
\begin{subfigure}  
\centering
(a)\includegraphics[scale=0.605]{DAG8BA}
\end{subfigure}
\quad 
\begin{subfigure} 
\centering
(b)\includegraphics[scale=0.57]{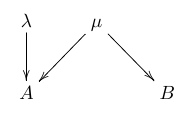}
\end{subfigure}
\caption{
(a) DAG for causal model defined by $A=\lambda\oplus B$ and $B=\nu$ (b) DAG for causal model defined by $A=\lambda\oplus\mu$ and $B=\mu$. 
}
\label{ExampleObsEqv} 
\end{figure}

An explicit example serves to illustrate this equivalence.  The causal model depicted in Fig.~\ref{ExampleObsEqv}(a), with functional dependences $A =\lambda \oplus B$ and $B = \nu$, involving a directed causal influence from $B$ to $A$,  is observationally equivalent to the causal model depicted in Fig.~\ref{ExampleObsEqv}(b), with functional dependences $A=\lambda \oplus \mu$ and $B=\mu$, which is purely common-cause.  To see this, note that one can express the functional dependences of the first causal model
 as $A=g(\lambda, \nu, B)=\lambda\oplus B$ and $B=f(\lambda, \nu)=\nu.$ Performing the substitution described in the previous paragraph yields $A=g(\lambda, \nu, f(\lambda, \nu))=g'(\lambda, \nu)=\lambda\oplus\nu$, which on identifying $\nu$ with $\mu$, results in the second causal model.

\subsection{Sufficiency of considering models with binary latents} \label{binary-latents}

We call a causal model where all the latent variables are binary a {\em causal model with binary latents}.  If there are $n$ binary latent variables, it is called an {\em $n$-latent-bit causal model}.  


\begin{thm}\label{thm:binarylatents}
Consider the family of causal models where the latent variables are discrete and finite, but not necessarily binary. Every such model is observationally equivalent to one with binary latents.  Equivalently, there is a causal model with binary latents in each observational equivalence class. 
\end{thm} 

The proof is provided in appendix~\ref{sufficientcy}, but we now present a simple example which illustrates the main idea of the proof.  

\begin{figure}[t] 
\begin{centering}  
\begin{subfigure}
(a)\includegraphics[scale=0.45]{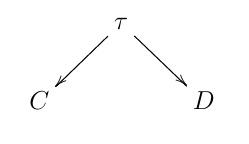}  
\end{subfigure}
\begin{subfigure} 
\centering
(b) \includegraphics[scale=0.58]{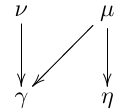}
\end{subfigure}
\\
\begin{subfigure} 
\centering
(c) \includegraphics[,scale=0.57]{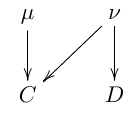}
\end{subfigure}
\caption{\color{black} Example of how to reduce a causal model with a latent trit to one involving only latent bits. (a) The original causal model, with functional dependences $C=\tau\mod2$, and $D=\big(2 (\tau\oplus_3 1)\big)\mod2$. (b) the trit $\tau$ is replaced by two bits, $\gamma$ and $\eta$,  which are presumed to be determined by a causal model having the depicted causal structure with functional dependences $\gamma = \mu\nu$, and $\eta =\mu$. (c) The causal model with binary latents that simulates the original model; the functional dependences are $C=\nu\mu \oplus_2 \nu$, and $D=\nu$. \color{black}}
\label{appendix fig 3}

\end{centering}
\end{figure}

Consider the causal model of Fig.~\ref{appendix fig 3}(a), where $C,D$ are binary, but $\tau$ is a three-valued variable, i.e., a trit. Suppose the functional relationships are as follows: $C=\tau\mod2$ and 
$D=\big(2 (\tau\oplus_3 1)\big)\mod2$,
 where $\oplus_k$ means addition modulo $k$. The values of $C,D$ for different values of $\tau$ are given in the table below.
\begin{center}
    \begin{tabular}{ | l | l | l | l |}
    \hline 
    $\tau$ & $C$  & $D$ \\ \hline 
    $0$ & $0$ & $0$ \\ \hline
    $1$ & $1$ & $1$  \\ \hline
    $2$ & $0$ & $1$  \\ \hline
        \end{tabular}
\end{center}
One can see that the distributions over $C,D$ that can be generated by this model correspond to the face of the tetrahedron that contains the vertices $\{[00], [11], [01]\}$. 

The trick to simulating this model using a $2$-latent-bit model is to replace the latent three-valued variable $\tau$ with a pair of binary variables $\gamma$ and $\eta$ and to imagine that these are causally related in the manner depicted in Fig.~\ref{appendix fig 3}(b).   
That is, we imagine a latent bit $\nu$ acting locally on $\gamma$ and a latent bit $\mu$ acting as a common cause of $\gamma$ and $\eta$ with the functional dependence $\gamma = \mu\nu$ and $\eta =\mu$. This causal model can generate any distribution over $\gamma$ and $\eta$ that has support only on the values $(\gamma,\eta)\in \{(0,1),(0,0),(1,1)\}$, \color{black} as can be seen by consulting the row containing class $(2,1,b)_{\rm{Id}}$ from the $3$-page table appearing later in this paper, where $A$ and $B$ play the role of $\gamma$ and $\eta$. \color{black}

If we take $\gamma$ and $\eta$ to be related to $\tau$ by $\tau = (\gamma \mod3) \oplus_3 (\eta \mod3)$, so that the values  $(0,0),(0,1)$ and $(1,1)$ of $(\gamma,\eta)$ map respectively to the values $0,1$ and $2$ of $\tau$, then any distribution over $\tau$ can be emulated by some distribution over the values  $(0,0),(0,1)$ and $(1,1)$ of $(\gamma,\eta)$ and hence some distribution over $\mu$ and $\nu$.
Finally, we can express $C$ and $D$ explicitly in terms of $\mu$ and $\nu$ by eliminating $\gamma$ and $\eta$, obtaining the causal model depicted in Fig.~\ref{appendix fig 3}(c) with dependences
$C=\nu\mu \oplus_2 \nu$ and $D=\nu$. 
By construction, we must obtain precisely the same semi-algebraic set for $C$ and $D$ in the model of Fig.~\ref{appendix fig 3}(c) as one does in the model of Fig.~\ref{appendix fig 3}(a).  We have therefore defined a $2$-latent-bit model that simulates our latent trit model. 

\color{black}
The key ingredient of the above example was that we were able find a causal model which could---by appropriately varying over the distribution of its latent variables---generate any distribution over a given face of the tetrahedron, and hence any distribution on a trit. In the case of an $m$-valued latent variable however, one would need to find a $k$-latent-bit model which could generate any distribution on an $m$-simplex. We provide an inductive procedure for constructing such a latent-bit model in appendix~\ref{sufficientcy}.
 
\color{black}

Theorem~\ref{thm:binarylatents} implies that for the project of determining the observational equivalence classes, it suffices 
 to consider models with binary latents. and  so we restrict our attention to these henceforth.
 
 \subsection{Inductive scheme}

Next, we define a scheme for composing pairs of $n$-latent-bit causal models into a single $(n+1)$-latent bit causal model, such that if we start with {\em all} possible pairs of $n$-latent-bit causal models, and apply the composition operation, we generate all possible $(n+1)$-latent-bit causal models. 

Denote the $n$ latent binary variables by $\lambda \equiv (\lambda_1, \dots, \lambda_n)$. A general $n$-latent-bit causal model is then defined by the functional dependences 
\begin{equation}\label{nlatentbitmodel}
A = \sum_{\alpha} a_{\alpha} \lambda^{\alpha} \quad\mathrm{and}\quad B = \sum_{\alpha} b_{\alpha} \lambda^{\alpha}
\end{equation}
 where $\lambda^{\alpha}$ is shorthand for the monomial $\lambda_1^{\alpha_1} \dots \lambda_n^{\alpha_n}$ for some set of exponents $\alpha \equiv (\alpha_1, \dots \alpha_n)$, and $a_{\alpha}, b_{\alpha} \in  \mathbb{Z}_2$ are parameters that specify the nature of the functional dependences. 

We assume that 
the first causal model is defined by parameters $\{a^{(0)}_{\alpha}\}$ and $\{b^{(0)}_{\alpha}\}$, and the second is defined by parameters $\{a^{(1)}_{\alpha}\}$ and $\{b^{(1)}_{\alpha}\}$.
The additional binary latent variable, which supplements the $n$ binary variables of the original two models is denoted $\delta$.  
The $(n+1)$-latent-bit model which is the composition of the two models is defined by the functional dependences 
\begin{align}\label{n+1latentbitmodel}
A = \sum_{\alpha} [(\delta \oplus 1) a^{(0)}_{\alpha} + \delta a^{(1)}_{\alpha}] \lambda^{\alpha},\nonumber\\
 B = \sum_{\alpha} [(\delta \oplus 1) b^{(0)}_{\alpha} + \delta b^{(1)}_{\alpha}] \lambda^{\alpha}.
\end{align}
This construction has been chosen such that $\delta$ acts as a switch variable: if we set $\delta=0$ in the resulting $(n+1)$-latent-bit model, we recover the first $n$-latent-bit model, while if we set $\delta=1$, we recover the second $n$-latent-bit model. 

\color{black}
With these definitions, our composition result can be summarized as follows. 
\begin{thm}
Consider the map that takes a pair of $n$-latent-bit causal models defined by the functional dependences of Eq.~(\ref{nlatentbitmodel}) with parameters $\{a^{(0)}_{\alpha}\} \cup \{b^{(0)}_{\alpha}\}$ for the first model, and parameters $\{a^{(1)}_{\alpha}\} \cup \{b^{(1)}_{\alpha}\}$ for the second model, and returns the $(n+1)$-latent-bit causal model defined by the functional dependences of Eq.~(\ref{n+1latentbitmodel}).  Under this map, the image of the set of all pairs of $n$-latent-bit causal models is the set of all $(n+1)$-latent-bit causal models.  
\end{thm}
\color{black}

\begin{proof}
The functional dependences of Eq.~(\ref{n+1latentbitmodel})
 can equivalently be expressed as polynomials in $\lambda$ and $\delta$ as 
\begin{align}
&A= \sum_{\alpha} \left( a^{(0)}_{\alpha} \lambda^{\alpha} +  ( a^{(0)}_{\alpha} \oplus a^{(1)}_{\alpha}) \lambda^{\alpha} \delta \right),\nonumber\\
&B= \sum_{\alpha} \left( b^{(0)}_{\alpha} \lambda^{\alpha} +  ( b^{(0)}_{\alpha} \oplus b^{(1)}_{\alpha}) \lambda^{\alpha} \delta \right)
\end{align}

It now suffices to note that as one varies over all possible joint values for the variables in the set $\{a^{(0)}_{\alpha}\} \cup \{a^{(1)}_{\alpha}\}$ (there are $2^{2^{n+1}}$ possibilities), one necessarily varies over all possible joint values for the variables in the set $\{a^{(0)}_{\alpha}\} \cup \{a^{(0)}_{\alpha} \oplus a^{(1)}_{\alpha}\}$, which in turn implies that one is varying over all possible polynomials in $\lambda_1, \dots, \lambda_n$ and $\delta$ in the expresson for $A$.  
By a similar argument, as one varies over all possible joint values for the variables in the set $\{b^{(0)}_{\alpha}\} \cup \{b^{(1)}_{\alpha}\}$, one varies over all possible polynomials in $\lambda_1, \dots, \lambda_n$ and $\delta$ in the expression for $B$.  It follows that as one varies over all possible joint values for the variables in the set $\{a^{(0)}_{\alpha}\} \cup \{a^{(1)}_{\alpha}\}\cup \{b^{(0)}_{\alpha}\} \cup \{b^{(1)}_{\alpha}\}$, one obtains all possible manners in which $A$ and $B$ might be functionally dependent on the latent variables in the $(n+1)$-latent-bit causal model.  Thus as one varies over all possible {\em pairs} of $n$-latent-bit causal models in our switch-variable construction, one varies over all possible $(n+1)$-latent-bit causal models. 
\end{proof}



We can therefore generate all causal models with binary latents by this inductive rule starting from the $0$-latent-bit causal models.


\subsection{Catalogue of observational equivalence classes}

Recall that two causal models are observationally equivalent if they define the same semi-algebraic set.  Thus, to characterize the observational equivalence classes, we proceed as follows.  For each new causal model that we generate by the inductive scheme, we determine the corresponding semi-algebraic set.  Every time one obtains a variety that has not appeared previously, one adds it to the catalogue of observational equivalence classes. 

Note that if a causal model has been obtained from two simpler models via our composition scheme, then the semi-algebraic set associated to it necessarily includes as subsets both of the semi-algebraic sets of the simpler models (note that this semi-algebraic set is generally {\em not} the convex hull of the semi-algebraic sets of the two simpler models).
It follows that if the semi-algebraic set of a given causal model is found to be the entire tetrahedron, then composing this model with any other will also yield the tetrahedron.  In this case, there are no new observational equivalence classes to be found among the descendants of this causal model in the inductive scheme.

In particular, if it were to occur that at some level of the inductive scheme, every newly generated causal model could be shown either to reduce to a previously generated causal model or to yield a semi-algebraic set that is the entire tetrahedron, 
then one could conclude that one's catalogue of the observational equivalence classes of causal models was complete in the sense that any $n$-latent bit causal model belongs to one of these classes.



We have used our inductive scheme to construct all observational equivalence classes generated by causal models with four or fewer binary latent variables.  We have also considered a large number of causal models with five binary latent variables and found no new observational equivalence classes.   This suggests that our catalogue may already be complete, although we do not have a proof of this.  Above, we noted circumstances in which our inductive scheme would terminate, which provides one strategy for attempting to settle the question.
Even in the absence of a proof of completeness, the inductive scheme presented here for classifying observational equivalence classes may be of independent interest to researchers in the field.


The observational equivalence classes of causal models that we have obtained (which cover all causal models with four or fewer binary latent variables) are presented in the table covering the next three pages.  For each class, we depict the semi-algebraic set that defines the class, the feasibility test for the class, and a representative causal model from the class.   Note that the open-interval constraints $0< p_{00},p_{01},p_{10},p_{11}<1$ are part of every feasibility test unless explicitly stated otherwise.  The corresponding constraint on the affine varieties is that those varieties confined to the edges exclude the vertices, those confined to the faces exclude the edges, and those in the bulk exclude the faces.

The task of describing the catalogue
 is simplified by the fact that many of the observational equivalence classes are related to one another by simple symmetries.  
We therefore organize the classes into orbits, where an orbit is a set of classes whose elements are related to one another by a set of symmetry transformations.  For one of the classes in the orbit \color{black}(which we term the `fiducial' class)\color{black}, we provide a full description, and below this description, we specify the set of symmetry transformations that must be applied to it to obtain the other elements of the orbit.  
 Formally, this is a set of representatives of the {\em right cosets} of the subgroup of symmetries of the semi-algebraic set in the full symmetry group of the tetrahedron. 



We express these representatives 
as compositions of the following set of 
 symmetry transformations, which we define below:
$\{\mathrm{Id}, f_A, f_B, S, X\}$.  For each of the five, \color{black}
we specify both their action on the causal model, i.e., their action on the functional dependences, from which their action on the DAG can be inferred, and on the elements of the joint distribution $\{p_{ab}:a,b\in \mathbb{Z}_2\}$, from which their action on the feasibility test can be inferred. Each symmetry transformation also defines an action on the tetrahedron in an obvious manner. Id is the identity transformation, leaving the model and $p_{ab}$ invariant; $f_A$ is the bit flip on $A$, replacing the functional dependence $A=f(\lambda)$ with $A=f(\lambda)\oplus 1$ and mapping $p_{ab} \to p_{a\oplus 1,b}$; $f_B$ is the bit flip on $B$, defined analogously to $f_A$; $S$ is the swap transformation, replacing the functional dependences $A=f(\lambda), B=g(\lambda)$ with $A=g(\lambda), B=f(\lambda)$, and mapping $p_{ab} \to p_{ba}$; $X$ is the ``add $B$ to $A$'' transformation, replacing the functional dependences $A=f(\lambda), B=g(\lambda)$ with $A=f(\lambda)\oplus g(\lambda), B=g(\lambda)$ and mapping $p_{ab}\to p_{a\oplus b,b}$. We denote a composition of two symmetry transformations by a right-to-left product: for instance, a bit flip on $A$ followed by a swap is denoted $Sf_A$.  The conjunction of a bit flip on $A$ and a bit flip on $B$ yields the same transformation regardless of the order in which they are implemented and is denoted $f_{AB}$. 

\color{black}
Finally, a given observational equivalence class will be distinguished by a label of the form  $(n,m,x)_g$.  Here, $n$ is the number of binary latent variables in the causal model, $m$ is the number of these that act as common causes, $x$ is an optional label that is used for distinguishing functional dependences that are consistent with a given $(n,m)$  but are observationally inequivalent, and $g$ labels the symmetry transformation that relates the class to the fiducial class

\onecolumn

\begin{centering}    
\begin{table}[h!]
  \begin{tabular}{| c | c | c | c | } 
    \hline
    \color{black} Class \color{black} & Semi-algebraic set & Test for feasibility & Minimal causal model \\ 
     && \color{black} $0 < p_{00},p_{01},p_{10},p_{11} < 1$ \color{black}&\\
     && \color{black} unless stated otherwise \color{black} &\\
    \hline   
   
$\color{black} (0,0)_{\rm Id} \color{black}$
   &
    \begin{minipage}{.3\textwidth}  
    \centering
      {\includegraphics[width=30mm, height=28mm]{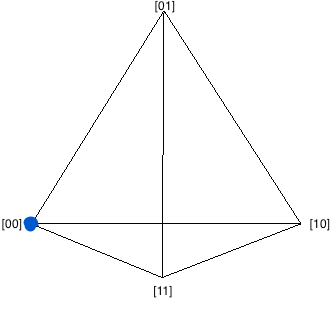}}
       \end{minipage}
   
    & 
    $\begin{aligned}  p_{00}&=1\\
    p_{01}=p_{10}&=p_{11}=0 \end{aligned}$  
    
     &
    \begin{minipage}{.3\textwidth}
    \centering
      {\includegraphics[width=28mm, height=20mm]{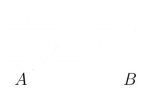} 
      \caption*{$A=0, \quad B=0$}}  
    \end{minipage} \\ \hline 
    
    \multicolumn{4}{|c|}{
   Transformations:  \color{black}$ G(0,0) = \{ {\rm Id}, \ f_A, \ f_B, \ f_{AB}\}$\color{black}} \\ \hline  
    
$\color{black} (1,0)_{\rm Id} \color{black}$
   &
    \begin{minipage}{.3\textwidth}  
    \centering
      {\includegraphics[width=33mm, height=31mm]{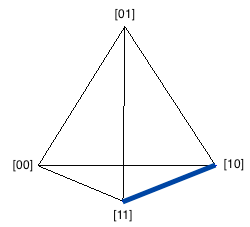}}
       \end{minipage}
    
    & 
    $\begin{aligned}  p_{00}=p_{01}=0 \end{aligned}$  
    &
    \begin{minipage}{.3\textwidth}
    \centering
      {\includegraphics[width=27mm, height=20mm]{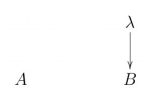} 
      \caption*{$A=1, \quad B=\lambda$}}  
    \end{minipage}
    
    \\ \hline
    
    \multicolumn{4}{|c|}{\color{black}$G(1,0)= \{ {\rm Id},  \ f_A, \  S, \ f_B S\} $\color{black}} \\ \hline  
    
$\color{black} (1,1)_{\rm Id} \color{black}$
      &
    \begin{minipage}{.3\textwidth}  
    \centering
      {\includegraphics[width=32mm, height=30mm]{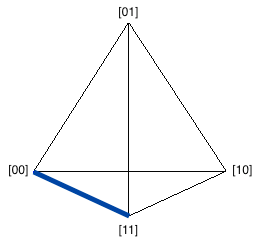}}
       \end{minipage}
    
    & 
    $\begin{aligned} p_{10}=p_{01}=0 \end{aligned}$  
    &
    \begin{minipage}{.3\textwidth}
    \centering
      {\includegraphics[width=28mm, height=22mm]{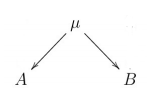} 
      \caption*{$A=\mu, \quad B=\mu$}}    
    \end{minipage}
    
    \\ \hline
    
    \multicolumn{4}{|c|}{\color{black}$G(1,1)=\{ {\rm Id}, \ f_A\}$\color{black}} \\ \hline   

$\color{black} (2,0)_{\rm Id} \color{black}$
   &
    \begin{minipage}{.3\textwidth} 
    \centering
      {\includegraphics[width=36mm, height=30mm]{AB}}
       \end{minipage}
    
    & 
    ${p_{00}p_{11}=p_{01}p_{10}}$  
    
    &
    \begin{minipage}{.3\textwidth}
    \centering
      {\includegraphics[width=30mm, height=22mm]{DAG8AB} 
      \caption*{$A=\lambda, \quad B=\nu$}}  
    \end{minipage}
    \\ \hline
    
    \multicolumn{4}{|c|}{\color{black} $G(2,0)=\{{\rm Id}\}$ \color{black}} \\ \hline  

$\color{black} (2,1,a)_{\rm Id} \color{black}$

&
    \begin{minipage}{.3\textwidth} 
    \centering
      {\includegraphics[width=36mm, height=30mm]{BcauseAFan}}
       \end{minipage}
    
    & 
    ${p_{00}p_{01}=p_{11}p_{10}}$  
    &
    \begin{minipage}{.3\textwidth}
    \centering
      {\includegraphics[width=30mm, height=22mm]{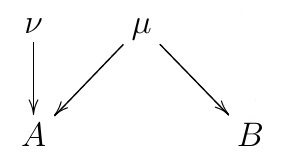} 
      \caption*{$A=\mu \oplus\nu, \quad B=\mu$}}  
    \end{minipage}
    
    \\ \hline
    
    \multicolumn{4}{|c|}{\color{black} $G(2,1,a)=\{ {\rm Id}, \ S \}$\color{black}} \\ \hline

$\color{black} (2,1,b)_{\rm Id} \color{black}$    
      &
    \begin{minipage}{.3\textwidth}  
    \centering
      {\includegraphics[width=36mm, height=37mm]{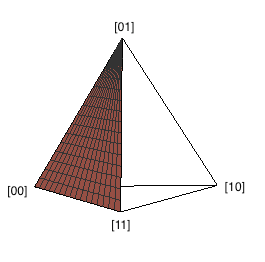}}
       \end{minipage}
    
    & 
    $\begin{aligned}  p_{10}=0 \end{aligned}$  
   &
    \begin{minipage}{.3\textwidth}
    \centering
      {\includegraphics[width=30mm, height=22mm]{DAG10} 
      \caption*{$A=\nu\mu, \quad B=\mu$}}    
    \end{minipage}
   
    \\ \hline
    
    \multicolumn{4}{|c|}{$\color{black} G(2,1,b)=\{\rm{Id}, \ f_A, \ f_B, \ S\}\color{black}$} \\ \hline     

$\color{black} (2,2)_{\rm{Id}} \color{black}$        
&
  \begin{minipage}{.3\textwidth}
    \centering
      {\includegraphics[width=39mm, height=30.5mm]{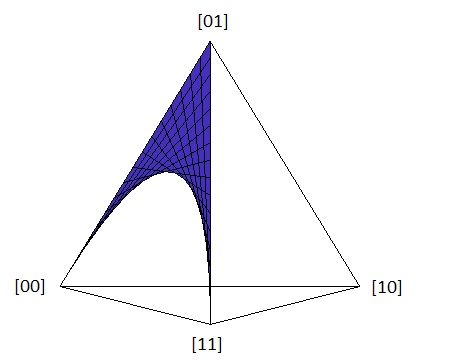}}
       \end{minipage}  
 
  &
    $\begin{aligned} (p_{01}+2p_{11}&-2)^2\geq{4}p_{00}, \\
    p_{10}&=0
    \end{aligned}$ 
    
    &
 \begin{minipage}{.3\textwidth}
    \centering
      {\includegraphics[width=37mm, height=27mm]{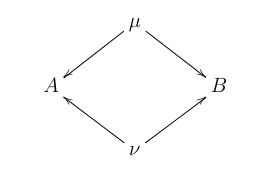} 
      \caption*{$A=\mu\nu,\quad B=\mu\oplus\nu\oplus\mu\nu$}}
       \end{minipage}   
    
    \\ \hline
    
  \multicolumn{4}{|c|}{$\begin{aligned}G(2,2)=\{\rm{Id}, \ S, \  f_A, \ f_B, \  f_{AB}X, \ f_BSX, \ f_A S {X}, \ {f_{AB}}S{X}, \ S{f_B}{X}S, \ f_{AB}S, \ f{X}S, \ {f_B}{X}S\end{aligned}\}$}  \\ \hline

    \end{tabular}
    \end{table}
 
    \begin{table}[h!]
  \begin{tabular}{| c | c | c | c | }
    \hline
   Class & Semi-algebraic set  & Test for feasibility & Minimal causal model  \\ \hline 

$\color{black} (3,1,a)_{\rm{Id}} \color{black}$      
&
  \begin{minipage}{.3\textwidth}
    \centering
     { \includegraphics[width=40mm, height=34mm]{Mult}}
       \end{minipage}  
 
  &
    $p_{00}p_{11}>{p_{01}p_{10}}$ 
    
    &
 \begin{minipage}{.3\textwidth}
    \centering
      {\includegraphics[width=30mm, height=18mm]{DAG} 
      \caption*{$A=\mu\nu, \quad B=\mu\lambda$}}
       \end{minipage}   
    
    \\ \hline 
    
  \multicolumn{4}{|c|}{ $G(3,1,a)=\{\rm{ Id}, \ f_A\} $}   \\ \hline      
    
    $\color{black} (3,1,b)_{\rm{Id}} \color{black}$    
&
  \begin{minipage}{.3\textwidth}
    \centering
      {\includegraphics[width=39mm, height=32mm]{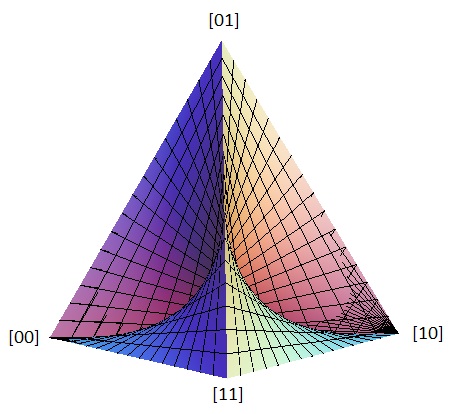}}
       \end{minipage}  
   
  &
   $\begin{aligned}(p_{01}-p_{11})(p_{00}p_{11}-{p_{01}p_{10}})&< 0 \\ (p_{01}-p_{11})(p_{10}p_{11}-{p_{01}p_{00}})&< 0 \end{aligned}$ 
     &
 \begin{minipage}{.3\textwidth}
    \centering
     { \includegraphics[width=30mm, height=18mm]{DAG} 
      \caption*{$A=\mu\oplus\nu, \quad B=\mu\lambda$}}  
       \end{minipage}
    
    \\ \hline
  
   \multicolumn{4}{|c|}{$G(3,1,b)=\{\rm{Id},  \ f_{AB} \}$} \\ \hline

$\color{black} (3,1,c)_{\rm{Id}} \color{black}$    
&
  \begin{minipage}{.3\textwidth} 
    \centering
      {\includegraphics[width=41mm, height=37mm]{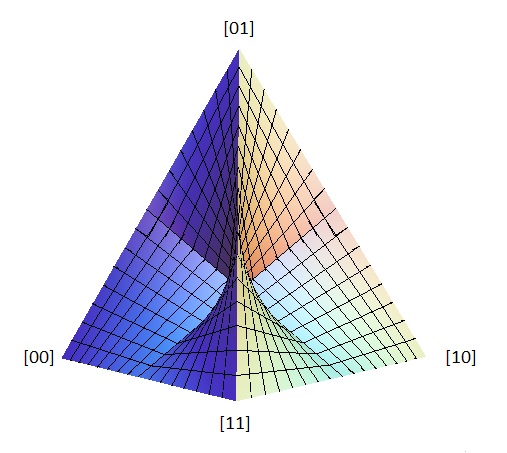}}
       \end{minipage}  
   
  &
   
   $\begin{aligned}
\frac{1}{4}&>\frac{p_{10}p_{11}-p_{00}p_{01}}{2p_{10}+p_{11}-1},\\
\frac{1}{4}&>\frac{p_{01}p_{11}-p_{00}p_{10}}{2p_{01}+p_{11}-1},\\
\frac{1}{4}&>\frac{p_{10}p_{01}-p_{00}p_{11}}{2p_{10}+p_{01}-1}.
\end{aligned}$ 

&
 \begin{minipage}{.3\textwidth}
    \centering
     { \includegraphics[width=30mm, height=18mm]{DAG} 
      \caption*{$A=\mu\oplus\nu, \quad B=\mu\oplus\lambda$}}
       \end{minipage} 

    \\ \hline
  
  \multicolumn{4}{|c|}{$G(3,1,c)=\{\rm{Id} \}$} \\ \hline

    $ (3,2,a)_{\rm{Id}}$  
&
  \begin{minipage}{.3\textwidth}
    \centering
     { \includegraphics[width=41.5mm, height=38mm]{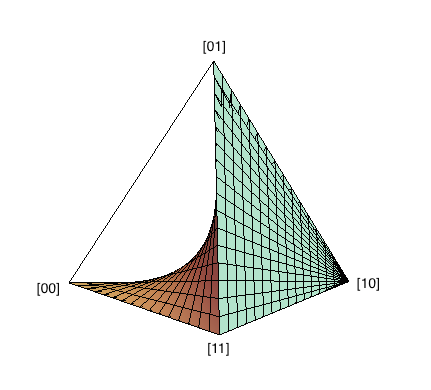}}
       \end{minipage}  

  &
    $p_{10}p_{11}>{p_{01}p_{00}}$ 
     &
 \begin{minipage}{.3\textwidth}
    \centering
      {\includegraphics[width=39mm, height=29mm]{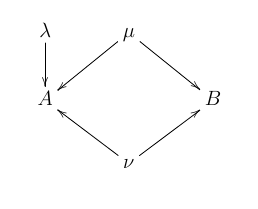} 
      \caption*{$A=\mu\nu\oplus\mu\lambda\oplus{1}, \quad B=\mu\nu\oplus{1}$}}
       \end{minipage}   
    \\ \hline 
    
  \multicolumn{4}{|c|}{$G(3,2,a)=\{ Id,  \ S{f_A}, \ f_A, \ f_A{X}{S}$\} }   \\ \hline

    $ (3,2,b)_{\rm{Id}}$  
&
  \begin{minipage}{.3\textwidth}
    \centering
      {\includegraphics[width=40mm, height=35mm]{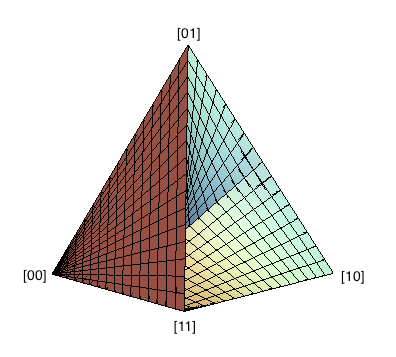}}
       \end{minipage}  
   
  &
   $\begin{aligned}(p_{01}-p_{10})(p_{11}p_{10}-{p_{01}p_{00}})&< 0 \\ (p_{01}-p_{10})(p_{00}p_{10}-{p_{01}p_{11}})&< 0 \end{aligned}$ 
    &
 \begin{minipage}{.3\textwidth}
    \centering
     { \includegraphics[width=38mm, height=28mm]{DAGleft} 
      \caption*{$A=\mu\oplus\nu\oplus\mu\lambda\oplus{1}, \quad B=\mu\oplus\nu\oplus{1}$}}
       \end{minipage} 
    
    \\ \hline
  
   \multicolumn{4}{|c|}{$G(3,2,b)=\{ Id,  \ f_{AB}{X}{S}, \ f_B{X}{S}, \ f_A\}$} \\ \hline 
   
       $ (3,2,c)_{\rm{Id}}$   
&
  \begin{minipage}{.3\textwidth}
    \centering
     { \includegraphics[width=42mm, height=35.5mm]{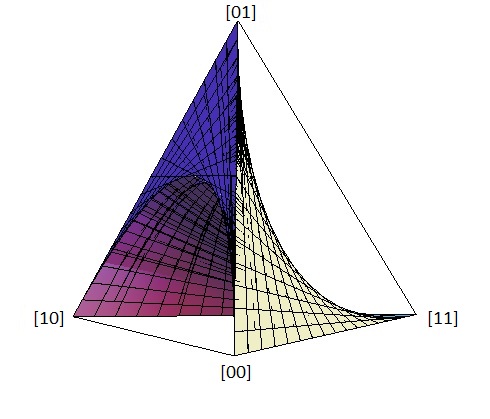}}
       \end{minipage}  
 
  &
   
   $\begin{aligned}
\vert {4}(p_{10}-&p_{11})(p_{00}p_{10}-p_{01}p_{11})\vert\leq \\ \big(p_{11}&(2p_{01}+2p_{10}+p_{00})- \\ &p_{10}(2p_{00}+2p_{11}+p_{01})\big)^2
\end{aligned}$ 
&
 \begin{minipage}{.3\textwidth}
    \centering
     { \includegraphics[width=35mm, height=29mm]{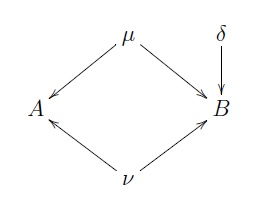} 
      \caption*{$A=\mu\nu, \quad B=\mu\oplus\nu\oplus\delta$}}
       \end{minipage}   
    \\ \hline   
    
  \multicolumn{4}{|c|}{ $\begin{aligned} G(3,2,c)=\{ \mathrm{Id}, \ S, f_A,  {f_{AB}}{S}{X}, \  f_B{X}{S}, \ {X}{S}, \ {f_{A}}{X}{S}, \ {f_{A}}{X}, \ {f_{AB}}{X}, \ f_A{S}{X}{S}, \ X \}\end{aligned}$} \\ \hline

\end{tabular} 
  
\end{table}

    \begin{table}[h!]
  \begin{tabular}{| c | c | c | c | }
    \hline
   Class & Semi-algebraic set  & Test for feasibility & Minimal causal model  \\ \hline    
 
 $(3,2,d)_{\rm{Id}}$
&
  \begin{minipage}{.3\textwidth}
    \begin{centering}
     { \includegraphics[width=40mm, height=35.5mm]{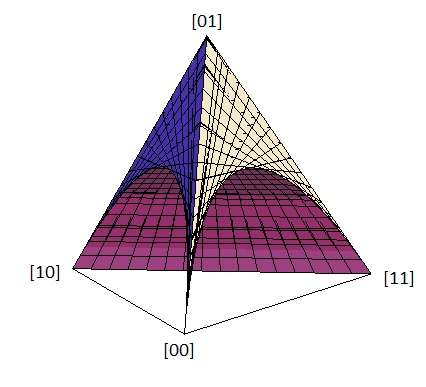}}\end{centering}
       \end{minipage}  

 &
   
$\big(p_{01}+2p_{10}+2p_{11}-2\big)^2\geq4p_{00}$
 &
 \begin{minipage}{.3\textwidth}
    \begin{centering}
     { \includegraphics[width=35mm, height=28mm]{DAGnew1} 
      \caption*{$A=\mu\oplus\nu, \quad B=\mu\nu\delta\oplus\mu\oplus\nu$}}\end{centering}
       \end{minipage}   

  \\ \hline
 
 \multicolumn{4}{|c|}{ $\begin{aligned}G(3,2,d)=\{ \mathrm{Id}, \ f_B, \ f_A, \ f_{AB}, \ {f_{AB}}{S}{X}, \ f_B{X}{S}, \ {X}{S}, \ {f_{A}}{X}{S}, \ {S}{X}, \ f_B{S}{X}, \  f_A{S}{X}, \ f_{AB}{S}{X}\} \end{aligned}$ } \\ \hline 
  
      $(3,2,e)_{\rm{Id}}$
&
  \begin{minipage}{.3\textwidth}
    \begin{centering}
     { \includegraphics[width=35mm, height=31.5mm]{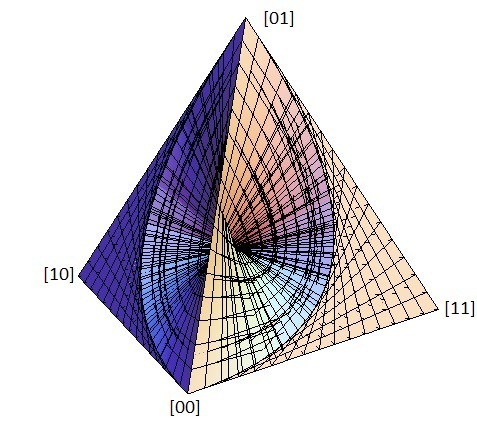}}\end{centering}
       \end{minipage}  
   
 &
   
$\begin{aligned}
{4}(p_{10}-p_{11}&)(p_{00}p_{10}-p_{01}p_{11})\leq \\ \big(p_{11}&(2p_{01}+p_{11})- \\ &p_{10}(2p_{00}+p_{10})\big)^2, \\
{4}(p_{10}-p_{11}&)(p_{01}p_{10}-p_{00}p_{11})\leq \\ \big(p_{11}&(2p_{01}+p_{11})- \\ &p_{10}(2p_{00}+p_{10})\big)^2
\end{aligned}$

&
 \begin{minipage}{.3\textwidth}
    \begin{centering}
      {\includegraphics[width=32mm, height=28.5mm]{DAGnew1} 
      \caption*{$A=\mu\oplus\nu, \quad B=\mu\nu\oplus\delta$}}\end{centering}
       \end{minipage} 
 \\ \hline 
  
 \multicolumn{4}{|c|}{$G(3,2,e)=\{\mathrm{Id}, \ f_{AB}{S}, \ f_{AB}{X}, \ f_B{X}, \ f_A{S}\}$ } \\ \hline

 $(3,2,f)_{\rm{Id}}$
&
  \begin{minipage}{.3\textwidth}
    \begin{centering}
     { \includegraphics[width=36mm, height=34mm]{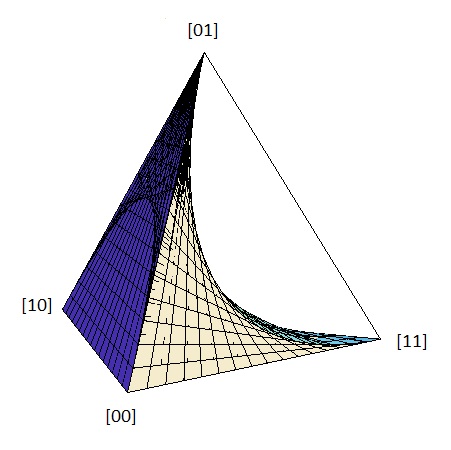}}\end{centering}
       \end{minipage}  
   
 &
   
$\begin{aligned}
\vert {4}(p_{10}&-p_{11})(p_{00}p_{10}-p_{01}p_{11})\vert\leq \\ \big(p_{11}&(2p_{01}+2p_{10}+p_{00})- \\ &p_{10}(2p_{00}+2p_{11}+p_{01})\big)^2 \\
&\\
p_{00}&p_{11}>  {p_{01}p_{10}}
\end{aligned}$

&
 \begin{minipage}{.3\textwidth}
    \begin{centering}
     { \includegraphics[width=32mm, height=27mm]{DAGnew1} 
      \caption*{$A=\nu\mu, \quad B=\mu\oplus\nu\delta$}}\end{centering}
       \end{minipage}  
  \\ \hline    
  
  \multicolumn{4}{|c|}{$\begin{aligned} G(3,2,f)=\{ \mathrm{Id}, \ S, f_A,  {f_{AB}}{S}{X}, \  f_B{X}{S}, \ {X}{S}, \ {f_{A}}{X}{S}, \ {f_{A}}{X}, \ {f_{AB}}{X}, \ f_A{S}{X}{S}, \ X \}\end{aligned}$} \\ \hline 
  
  $(3,2,g)_{\rm{Id}}$
&
  \begin{minipage}{.3\textwidth}
    \centering
     { \includegraphics[width=36mm, height=32mm]{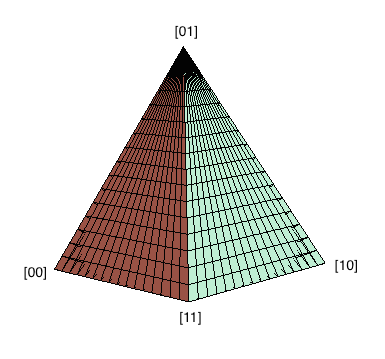}}
       \end{minipage}  
 
  &

&
 \begin{minipage}{.3\textwidth}
    \centering
     { \includegraphics[width=30mm, height=24mm]{DAGnew1} 
      \caption*{$A=\mu\nu\oplus 1, \quad B=\mu\nu\delta\oplus\nu$}}
       \end{minipage}   
    \\ \hline   
    
  \multicolumn{4}{|c|}{$G(3,2,g)=\{\rm{ Id}\}$} \\ \hline  
  $(3,3)_{\rm{Id}}$
&
  \begin{minipage}{.3\textwidth}
    \begin{centering}
     { \includegraphics[width=31mm, height=26mm]{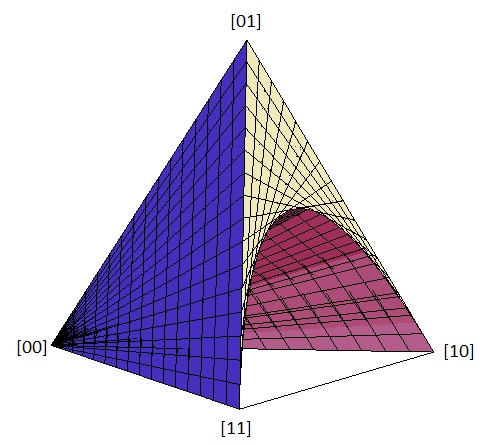}}\end{centering}
       \end{minipage}  
  
 &
   
$\big(2p_{10}+p_{01}\big)^2\geq4(1-p_{00})p_{10}$
&
 \begin{minipage}{.3\textwidth}
    \begin{centering}
     { \includegraphics[width=32.5mm, height=24mm]{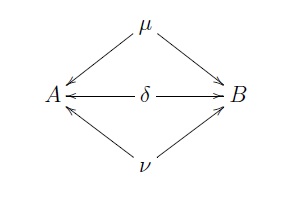} 
      \caption*{$A=\delta\mu\oplus\delta\nu\oplus\delta, \quad B=\delta\mu\nu\oplus\delta$}}\end{centering}
       \end{minipage}  

  \\ \hline   
  
  \multicolumn{4}{|c|}{$\begin{aligned}G(3,3)=\{ \mathrm{Id}, \ f_B, \ f_A, \ f_{AB}, \ {f_{AB}}{S}{X}, \  f_B{X}{S}, \ {X}{S}, \ {f_{A}}{X}{S}, \ {S}{X}, \ f_B{S}{X}, \  f_A{S}{X}, \ f_{AB}{S}{X}\} \end{aligned}$ } \\ \hline

  $(4,2,a)_{\rm{Id}}$
&
  \begin{minipage}{.3\textwidth}
    \begin{centering}
      {\includegraphics[width=33mm, height=28mm]{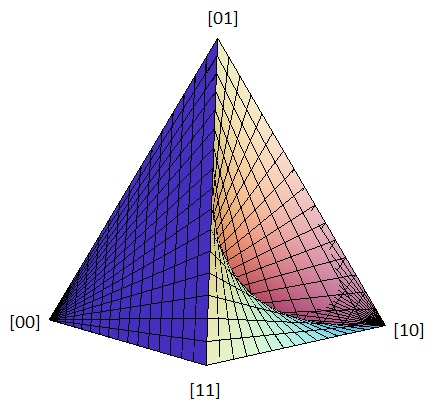}}\end{centering}
       \end{minipage}  
 
 &
   
$\begin{aligned} 
p_{00}p_{11}&>{p_{01}p_{10}}\\
p_{11}p_{10}&>{p_{00}p_{01}} \end{aligned}$
& 
 \begin{minipage}{.3\textwidth}
    \begin{centering}
      {\includegraphics[width=33mm, height=20mm]{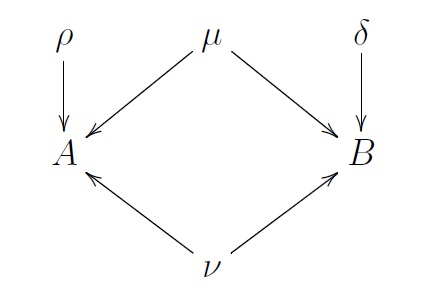} 
      \caption*{$A=\nu\mu\oplus\nu\rho, \quad B=\nu\mu\delta$}}\end{centering}
       \end{minipage}   

  \\ \hline 
  
  \multicolumn{4}{|c|}{  $\begin{aligned} G(4,2,a)=\{ \mathrm{Id}, \ f_B, \ f_A, \ f_{AB}, \ {f_{AB}}{S}X{f_A}, \  f_B X{S}{f_A}, \ {X}{S}, \ {f_{A}}X{S}{f_A}, \ {S}X, \ f_B{S}X{f_{AB}}, \  f_{AB}X{S}{f_B}, \ f_{A}{S}X{f_B}\} \end{aligned}$} \\ \hline

  $(4,2,b)_{\rm{Id}}$
&
  \begin{minipage}{.3\textwidth}
    \begin{centering}
      {\includegraphics[width=34mm, height=30mm]{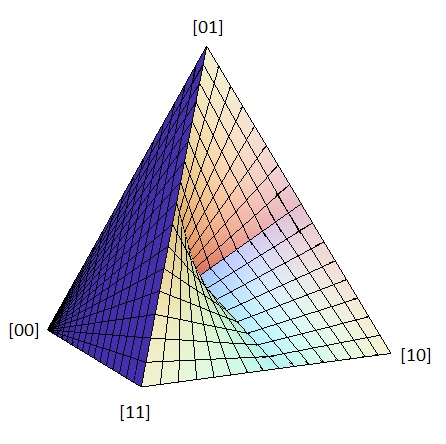}}\end{centering}
       \end{minipage}     
   
 &
   
$\begin{aligned}
p_{00}p_{11}&>{p_{01}p_{10}}\\
p_{11}p_{10}&>{p_{00}p_{01}} \\
{p_{11}p_{01}}&> p_{00}p_{10}
\end{aligned}$
&
 \begin{minipage}{.3\textwidth}
    \begin{centering}
     { \includegraphics[width=31mm, height=20mm]{DAGconvex} 
      \caption*{$A=\nu\mu\oplus\nu\rho, \quad B=\nu\mu\oplus\nu\delta$}}\end{centering}
       \end{minipage} 

  \\ \hline  
  
  \multicolumn{4}{|c|}{$G(4,2,b)=\{\mathrm{Id}, \ f_A, \ f_B, \ f_{AB} \}$} \\ \hline

   \end{tabular}
  
\end{table} 
\end{centering}

\twocolumn

for the particular $n, m$ and $x$. The set of possibilities for $g$ for a given $n, m,$ and $x$ is a subgroup of the group closure of $\{ Id, f_A, f_B, S,X\}$, which we denote by $G(n,m,x)$.
Note that $n,m \in \mathbb{N}$, $m\le n$, and 
we take $x \in \{ a, b, c, \dots \}$.  
\color{black}

The first few steps of our iterative procedure for the construction of causal models proceed as follows. 

The semi-algebraic sets associated to the four $0$-latent-bit causal models are the four vertices of the tetrahedron, labelled by the deterministic assignments to $A$ and $B$, that is, as $[00]$, $[10]$, $[01]$ and $[11]$.
\color{black} These correspond to the classes $\{ (0,0)_g : g\in \{ {\rm Id},f_A,f_B,f_{AB}\}\}$, depicted in the first row of the table (because there is only one observational equivalence class with $n=0$ and $m=0$, the label $x$ is not necessary in this case and so it is not excluded from the name of the class). \color{black} 

One finds that by composing these with one another into $1$-latent-bit causal models, one arrives at six new observational equivalence classes.  Four of these correspond to models with a single latent bit that acts locally, and their semi-algebraic sets are the four edges of the tetrahedron with endpoints $\{[00],[01]\}$, $\{[10],[11]\}$, $\{[00],[10]\}$, $\{[01],[11]\},$ which we might call the {\em $AB$-uncorrelated edges}; \color{black} these correspond to the classes $\{ (1,0)_g : g\in \{ {\rm Id},f_A,S,f_{B}S\}\}$, depicted in the second row of the table. \color{black} 
 Two of these correspond to models with a single latent bit acting as a common cause and their semi-algebraic sets are the $[00]$-$[11]$ and $[01]$-$[10]$  edges of the tetrahedron, which we might call the {\em $AB$-correlated edges}, \color{black} corresponding to the classes $\{(1,1)_g : g\in \{ {\rm Id},f_A\}\}$, depicted in the third row of the table. \color{black} 

 Next, one constructs all of the $2$-latent-bit causal models and finds their semi-algebraic sets. This set includes the model of Fig.~2(c), where both latent bits act locally, and whose semi-algebraic set is the fan of Fig.~2(d), which touches each of the $AB$-uncorrelated edges of the tetrahedron, \color{black} corresponding to the single class $(2,0)_{\rm Id}$, depicted in the fourth row of the table. \color{black} 
 This set also includes the models of Fig.~1(a) and Fig.~2(a) where one of the latent bits acts as a common cause and whose semi-algebraic sets are the fans of  Fig.~1(b) and Fig.~2(b), which touch the $AB$-correlated edges of the tetrahedron.  
 \color{black} They correspond to the pair of classes $\{ (2,1,a)_{g}: g\in \{ {\rm Id}, S\}\}$, depicted in the fifth row of the table. \color{black} 
There is also a second type of $2$-latent-bit causal model where one latent bit acts as a common cause 
which yield the semi-algebraic sets corresponding to the four faces of the tetrahedron.  \color{black} These are the four classes $\{ (2,1,b)_g: g\in \{ {\rm Id},f_A,f_B,S\} \}$ in the table.  \color{black}
 \color{black} When both of the latent variables act as common causes, one obtains semi-algebraic sets that are subsets of a face of the tetrahedron \color{black} and  
 which have the appearance of the StarFleet insignia from Star Trek, of which there are twelve in total.   \color{black} These are the classes labelled $(2,2)_g$
   in the table. \color{black}
   The construction of $3$-latent-bit and $4$-latent-bit causal models proceeds similarly and the new observational equivalence classes one thereby obtains are depicted in the rest of the table.
   
   \color{black}
   \section{Identification of parameters in the causal model}
   
Our results also have applications for the identification problem, that is, the problem of determining which parameters in a causal model can be identified or bounded using observational data.
   
Consider the problem of identifying the probability distributions over the latent variables (our $q_j$ parameters) in a causal model associated to a given functional causal structure.  From the description of our algorithm, it is clear that the $q_j$ parameters 
 are generally identifiable because the Groebner basis provides a means of computing expressions for them in terms of our $p_i$ parameters (the observational data).  Indeed, we often must compute the explicit expressions for one or more of the $q_j$s in terms of the $p_i$s {\em as an intermediate step} on the way to deriving our feasibility tests.  Eq.~\eqref{identification1} is an example of such an identification formula.

The other sort of parameter of a causal model that one may wish to identify is the nature of the functional dependences (assuming the model is indeed functional).   For the sorts of models we consider, this problem is also solved by our results. 

Consider the problem where the causal structure is given, but where there is uncertainty over the nature of the functional dependences thereon.  
For instance, suppose that it is known that the functional causal structure is {\em either} the minimal structure associated to the class $(2,1,a)_{\rm Id}$ in our table or the one associated to $(2,1,b)_{\rm Id}$. Because the semi-algebraic sets defining these two classes do not intersect\footnote{Recall our convention of demanding the probabilities for latent variable to be bounded away from 0 and 1, so that all of our semi-algebraic sets are confined to the interior of the tetrahedron.}, it is clear that one can settle this decision problem on the basis of the observational data.  

As another, more nuanced example, suppose that it is known that the functional causal structure is the minimal structure associated to one of the three classes $(3,1,a)_{\rm Id}$, $(3,1,b)_{\rm Id}$, and $(3,1,c)_{\rm Id}$  in our table.  Here, one finds that certain points in the space of distributions over the observed variables pass the feasibility test for just one of these functional causal structures, other points pass the test for two of them, while still others pass the test for all three.  

More generally, one might know {\em only} the causal structure.  For instance, the set of possible functional causal structures might be the minimal ones in each of the classes in the set $\{(2,1,a)_{g}:g \in \{\rm{Id},  \ S\}\} \cup \{(2,1,b)_{g}:g \in \{\rm{Id}, \ f_A, \ f_B, \ S\}\}$.  The feasibility tests we have derived provide a means of determining, for any given point in the space of distributions over the observed variables, precisely which of these functional causal structures is compatible with that observational data.  
\color{black}
   
    \section{Discussion}  \label{discussion}

\subsection{Future directions}

The restriction to pairs of binary observed variables is a limitation of our analysis. In future work, we hope to extend our approach to cases where the observed variables have an arbitrary number of values and where the number of observed variables is also arbitrary. While the tools from algebraic geometry employed in this paper provide a procedure for deriving feasibility tests for such functional causal structures in principle, in practice it is unlikely that such procedures will be scalable. Indeed, calculating Groebner bases is an $\mathrm{EXPSPACE}$-complete problem \cite{complexity}. Nevertheless, it may still be possible to develop new tools for causal inference in these cases using the approach described here.

It also remains an open problem to decide, for any given functional causal structure, which observational equivalence class it belongs to. That is,  even if our catalogue of classes is complete, it merely establishes that every functional causal structure falls into one of these classes, but it does not provide a means of deciding, for a given functional causal structure, having an arbitrary number of latent variables and functional dependences, which class it is a member of.  \color{black} Of course, if one supplements a given functional causal structure with distributions over the latent variables, then one obtains a joint distribution over the observed variables and this can be subjected to the feasibility tests for different observational equivalence classes. It is likely, however, that there are better ways of solving the classification problem, for instance, by determining how the functional dependences can be simplified.  \color{black} Solving this classification problem would allow one to find common features of all of the functional causal structures in a given class, for instance, features of the topology of the causal structure.

We have here made the idealization that the uncontrolled statistical data is given as a joint distribution whereas in practice it is a finite sample from this distribution. To contend with this idealization, one should in practice evaluate causal models by considering how well the finite statistical data can be fit to them.

\subsection{Relevance to quantum foundations}\label{quantum}

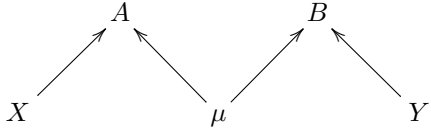
\begin{figure}[t]
\begin{center}
\begin{displaymath}
    \xymatrix{ &A&& B&  & \\
             X \ar[ur] && \ar[ul] \mu \ar[ur] && Y \ar[ul]\\
        }
\end{displaymath}
\end{center}
\caption{\color{black} In the Bell scenario, one is interested in the conditional distribution $\mathbb{P}(A,B|X,Y)$.  This is equivalent to a set of distributions over $A,B$, $\{\mathbb{P}_{(X,Y)}(A,B)\}$, one for each choice of measurement setting.\color{black}} \label{BellScenario}
\end{figure}

One of the motivations of the current work was the prospect of new insights into the interplay between causal structure and observed correlations in quantum theory. In particular, for a pair of quantum systems---each subjected to one of a set of possible measurements---a {\em Bell inequality} \cite{Bell, CHSH, BellReview} is a constraint on the joint probability distribution over the outcomes of each possible choice of the local measurements (that is, for every combination of local measurement {\em settings}).  It has recently been noted~\cite{RC, Fritz1} that one can understand the assumptions required to derive a Bell inequality as the standard assumptions for causal inference together with a particular hypothesis about the underlying causal structure, namely, that each local outcome depends causally on the corresponding local setting and on a latent common cause between the two systems. \color{black} This causal structure is illustrated in Fig.\ref{BellScenario}, where $A$ and $B$ are the measurement outcomes for each quantum system, $X$ and $Y$ are the local choices of measurement, and $\mu$ is the latent common cause. The complete set of Bell inequalities for this scenario, therefore, can be understood as a feasibility test for such a causal model. \color{black}
  
The problem considered \color{black} in the current work is \color{black} different from that of deriving the complete set of Bell inequalities in a couple of ways: (i) The observational input to our causal inference problem is different; there are no setting variables in our problem\color{black}---that is to say, any variable distinct from the observed $A,B$ appearing in a causal structure must be latent---\color{black}and therefore our input is a single joint distribution over two observed binary variables rather than a {\em set} of such distributions, one for each choice of the setting variables.
  (ii) The hypotheses whose feasibility we are testing are different; while the set of all Bell inequalities provides a test of the feasibility \color{black} of the causal structure illustrated in Fig.~\ref{BellScenario}\color{black}, we here seek to assess the feasibility of a causal structure for a given choice of cardinalities for the latent variables appearing therein (e.g., whether a given latent variable consists of a single bit, two bits, etcetera) and for a given choice of the precise form of the functional dependence of the observed variables on the latent variables.

\color{black} Consider the Bell scenario of Fig.~\ref{BellScenario} where $A,B,X,$ and $Y$ are binary variables.  To define a functional causal structure, one must supplement this causal structure with a hypothesis about the cardinality of $\mu$ and a hypothesis about the function $f$ that maps $X,\mu$ to $A$ and the function $g$ that maps $Y,\mu$ to $B$.  (Given that there are only 16 possible values of $A,B,X,$ and $Y$, a $\mu$ of cardinality 16 is sufficient to simulate any other case.)  The conditional distributions $\mathbb{P}(A,B|X,Y)$ compatible with this functional causal structure are
\begin{equation}\label{Belleq}
\begin{aligned}
\mathbb{P}(A,B|X,Y)&=\sum_\mu \mathbb{P}(A,B|X,Y,\mu)q_\mu \\
&=\sum_{\mu | A=f(X,\mu), B=g(Y,\mu)}q_\mu
\end{aligned}
\end{equation}
where $q_{\mu}$ denotes a probability distribution over the latent variable $\mu$.   To determine the semi-algebraic set of possibilities for $\mathbb{P}(A,B|X,Y)$ that are compatible with this functional causal structure, one could use the techniques of the present article.  From the polynomial equalities that hold between the $\mathbb{P}(A,B|X,Y)$ and the $q_{\mu}$ (those given in Eq.~\eqref{Belleq}), one seeks to obtain constraints on the $\mathbb{P}(A,B|X,Y)$ alone by eliminating the $q_{\mu}$.  Because the variables to be eliminated appear linearly, to eliminate the $q_{\mu}$, it suffices to use quantifier elimination techniques that are less computationally demanding than implicitization, such as Fourier-Motzkin elimination.

Note that if some observed correlations violate an inequality derived in this fashion, it only establishes the infeasibility of a given classical functional causal structure.  Violation of Bell inequalities, on the other hand, rule out the feasibility of the causal structure, regardless of the cardinality of the latent variables and the nature of the functional dependences.  In this sense, deriving Bell inequalities is more challenging than deriving feasibility tests for functional causal structures.  However, in another sense, deriving Bell inequalities is more straightforward because the semi-algebraic set defined by Eq.~\eqref{Belleq} is a polytope, whereas for a general funcational causal structure this is not the case. The mathematical tools that have been used to derive Bell-type inequalities---which include semi-definite~\cite{In2} and linear programming~\cite{In3} as well as Fourier-Motzkin elimination~\cite{Qalgebraic2, Qalgebraic1}---are therefore quite different from those used here.

\color{black}


Bell inequalities are significant to the foundations of quantum theory because they are found to be violated in experiments on pairs of separated quantum systems, implying that the predictions of quantum theory cannot be explained by \color{black} a classical causal model with \color{black} the causal structure that one expects to hold for the experiment (that of Fig.~\ref{BellScenario}) without fine-tuning~\cite{RC}.  

Researchers in the field of quantum foundations have now begun to apply insights obtained from the study of Bell inequalities to the problem of deriving constraints on observed correlations in more general causal scenarios~\cite{Fritz1, Fritz2, Chaves, chaves1, star, bilocal}, and the current work constitutes another contribution in this direction.

More importantly, there are now a few proposals for how to generalize the standard notion of a causal model to the quantum realm.  Ref.~\cite{LR}, for instance, proposes a definition of a quantum causal model in terms of a noncommutative generalization of conditional probability, while \color{black} Refs.~\cite{Fritz2,hlp, Caslav} follow a more operational approach. 
With a notion of quantum causal model in hand, one can explore the problem of inferring facts about the quantum causal model from observed correlations.  This is the problem of {\em quantum causal inference}. 

In the case of Bell-type experiments, for instance, one expects a quantum causal model with the natural causal structure (that of Fig.~\ref{BellScenario})
 to be feasible only if the observed correlations satisfy the so-called Cirel'son bound, which is a generalization of a Bell inequality \cite{Tsirelson}. A simple case of quantum causal inference that has been investigated recently is the problem of distinguishing a cause-effect relation from a common-cause relation.   
Here, it has been shown that the quantum correlations can distinguish the two cases even in uncontrolled experiments, implying a quantum advantage for causal inference~\cite{RK}.  

 \color{black}  In quantum causal models, variables are replaced by systems, each represented by a Hilbert space, and one makes a distinction between observed systems, upon which a measurement is made, and latent systems.  Sets of systems are described by joint quantum states (as opposed to the joint probability distributions that describe sets of variables), and the functional dependences are specified by unitary maps.  A natural analogue of the classical causal inference problem is to make inferences about the causal structure and the parameters of the causal model given a {\em joint quantum state} on the observed systems.  The natural analogue of the functional causal structures considered in this article are quantum causal structures together with a specification of the dimensions of the latent systems and the unitaries that describe the functional dependences.  To derive a feasibility test for a functional causal structure, one must eliminate the real-valued parameters that specify the quantum state of the latent systems.  For example, if a given latent system is 2-dimensional (the quantum analogue of a binary latent variable), there are {\em three} real-valued parameters needed to  specify the state completely (as opposed to the one real parameter needed to completely specify a distribution over a classical bit).  The expectation values of the three Pauli operators, for instance, suffice to do so.  Nonetheless, one can still take advantage of the techniques from algebraic geometry employed in this work to eliminate these parameters and determine constraints on the quantum state of the observed systems. In this way, we ought to be able to derive feasability tests for functional causal structures in the quantum sphere.
\color{black}
 



\subsection{Related work} \label{related}

\color{black} The extent to which the mathematical tools associated to quantifier elimination are well-suited to problems of causal inference has been previously emphasized by Geiger and Meek~\cite{Geiger-Meek99}.  
Many authors have noted, in particular, the applicability of quantifier elimination to the problem of deriving tests for the feasibility of a causal structure when the cardinality of the latent variables is known.   Ref.~\cite{Geiger-Meek99}, for instance, used Cylindrical Algebraic Decomposition to derive equality and inequality constraints for a particular causal model. However the computational complexity of such brute-force quantifier elimination (doubly exponential in the number of parameters) means that its applications are limited to very simple examples. 

Many previous works have appealed to implicitization procedures using Groebner bases to obtain equality constraints for causal models.
Geiger and Meek~\cite{Geiger-Meek}, Garcia, Stillman and Sturmfels \cite{Garcia2}, and Garcia \cite{Garcia} have used implicitization to obtain the smallest algebraic variety that contains the semi-algebraic set of joint distributions over observed variables for various causal structures with known cardinalities of latent variables.  This yields polynomial equalities on the joint distribution whose satisfaction are necessary conditions for compatibility with the causal structure.
Kang and Tian~\cite{intervension2} have also applied implicitization techniques to the problem of identifying polynomial equality constraints on observational and interventional distributions (using the framework supplied by Refs.~\cite{In1,intervension1}).



Our work goes beyond these treatments insofar as it uses implicitization as one step in an algorithm that finds the semi-algebraic set itself rather than the smallest algebraic variety containing it.  The second step is to use the extension theorem (described in Appendix A) to find inequality constraints on the joint probability distribution over observed variables from knowledge of the Groebner basis.  To illustrate the difference, consider the observational equivalence class labelled $(2,2)_{\rm{Id}}$ in our classification.  This corresponds to a semi-algebraic set for which the smallest algebraic variety containing it is the plane $p_{10}=0$.  The intersection of this variety with the tetrahedral probability simplex is its $p_{10}=0$ facet.  The semi-algebraic set, however, is a strict subset of this, the one satisfying the additional inequality $(p_{01}+2p_{11}-2)^2\geq{4}p_{00}$.

One novel feature of our approach which distinguishes it from previous uses of implicitization is that we focus on deriving feasibility constraints for a causal structure with specific functional dependences.  In previous approaches, the set of variables that needed to be eliminated included {\em both} the parameters describing the probability distributions for the root variables and the parameters describing the conditional probability distributions for each non-root variable.  In our approach, the second sort of parameter is fixed and not in need of elimination.  The restriction to binary variables ensures that the number of distinct possible functional dependences is relatively modest. 



Finally, the use of Groebner bases in identifying or bounding parameters in a causal model has also been highlighted in previous work such as Garcia-Puente {\em et al.}\cite{SSG}.

\color{black}  After the completion of this work, we became aware of related independent works by Chaves~\cite{chaves} and Rossett \emph{et al.}~\cite{gisin} which also derive nonlinear inequalities for determining the feasibility of certain causal structures.  These authors consider structures which, like Bell scenarios, have multiple pairs of observed variables that are related as cause and effect (understood as setting-outcome pairs) but which, unlike Bell scenarios, can have more than one latent common cause acting on the outcome variables. 
Chaves simplifies the quantifier elimination problem that must be solved using a round of Fourier-Motzkin elimination, while
Rossett \emph{et al.} provide an inductive approach for deriving new inequalities from given inequalities for subgraphs of the causal network under consideration. 
Combining our approach with these other methods  constitutes an interesting direction for future work.\color{black}

\color{black}

\vspace{-.37cm}
\section*{Acknowledgements}
This research began while CML was completing the Perimeter Scholars International Masters program at Perimeter Institute. Research at Perimeter Institute is supported by the Government of Canada through Industry Canada and by the Province of Ontario through the Ministry of Research and Innovation.

\onecolumn

\section*{Appendices} 
\begin{appendix}

\section{Ideals, varieties and Groebner bases} \label{alg geo}

We now introduce useful concepts and tools from algebraic geometry that we will make use of in solving the problem mentioned in section 2 of the main text. We will follow the presentation given in \cite{geo}.

We define a monomial in $x_1,\dots,x_n$ to be a product of the form $x_1^{\alpha_1}x_2^{\alpha_2}\dots{x_n^{\alpha_n}}$, where the exponents are non-negative integers, $\alpha_i\in\mathbb{Z}_{\geq{0}}$ for $i=1,\dots,n$. We can simplify our notation slightly by letting $\alpha=(\alpha_1,\dots,\alpha_n)$ and setting $$x^{\alpha}=x_1^{\alpha_1}x_2^{\alpha_2}\dots{x_n^{\alpha_n}}.$$ We can now define a polynomial over a field $k$.
\begin{define}
A polynomial $f$ in $x_1,\dots,x_n$ with coefficients in a field $k$ is a finite linear combination of monomials. We write $f$ as $$f=\sum_{\alpha}c_{\alpha}x^{\alpha},\quad{c_{\alpha}\in{k}},$$ where the sum is taken over a finite number of $\alpha$'s. 
\end{define}
The set of all polynomials in $x_1,\dots,x_n$ with coefficients in $k$ is denoted $k[x_1,\dots,x_n]$. When we deal with polynomials we are mainly interested in the solution set of systems of polynomial equations. This leads us to the main geometrical objects studied in algebraic geometry, algebraic varieties and semi-algebraic sets, which we now define.

\begin{define}
Let $k$ be a field and let $f_1,\dots,f_s$ be polynomials in $k[x_1,\dots,x_n]$. Then we set
$$\begin{aligned} \bold{V}(f_1,\dots,f_s)=\{(a_1,\dots,a_n)\in{k^n}: f_i(a_1,\dots,a_n)=0, \forall,{1\leq{i}\leq{s}}\}.\end{aligned}$$
We call $\bold{V}(f_1,\dots,f_s)$ the \emph{algebraic variety} (also called the \emph{affine variety}) defined by $f_1,\dots,f_s$.
\end{define}

Thus, an algebraic variety $\bold{V}(f_1,\dots,f_s)\subset{k^n}$ is the solution set of the system of polynomial equations $f_1(x_1,\dots,x_n)=\dots=f_s(x_1,\dots,x_n)=0.$ A \emph{basic semi-algebraic set} is defined to be the solution set of a system of polynomial equalities and \emph{in}equalities, that is:

\color{black}
\begin{define}
A \emph{basic semi-algebraic set} is defined by $\{x\in \mathbb{R}^n: g_i(x)\rightleftharpoons 0, \text{ }\forall \text{ } i=1,\dots, m\},$ where $g_1,\dots,g_n \in \mathbb{R}[x_1,\dots,x_n]$ are polynomials over the reals\footnote{Note that one can replace the real field $\mathbb{R}$ used in the last definition with any \emph{ordered} field.} and where $\rightleftharpoons \text{corresponds to either } \geq,\text{ } =, \text{ or } \leq$.  
\end{define}

Note that algebraic varieties are examples of basic semi-algebraic sets.  

 \begin{define}
A \emph{semi-algebraic set} is formed by taking finite combinations of unions, intersections, or complements of basic semi-algebraic sets. 
\end{define}

\color{black}

For {\em any} causal model, the set of possible joint distributions that can be generated by it are represented by a semi-algebraic set.   It follows that two causal models are observationally equivalent if and only if they generate the same semi-algebraic set.


We now introduce and define \emph{ideals}, the main algebraic object studied in algebraic geometry.
\begin{define}
A subset $I\subset{k[x_1,\dots,x_n]}$ is an \emph{ideal} if it satisfies: \begin{enumerate}
\item $0\in{I}$,
\item If $f,g\in{I}$, then $f+g\in{I}$,
\item If $f\in{I}$ and $h\in{k[x_1,\dots,x_n]}$, then $hf\in{I}$.
\end{enumerate}
\end{define}

A natural example of an ideal is the ideal generated by a finite number of polynomials.
\begin{define}
Let $f_1,\dots,f_s$ be polynomials in ${k[x_1,\dots,x_n]}$. Then we set 
$$\langle{f_1,\dots,f_s}\rangle=\Big\lbrace\sum_{i=1}^sh_if_i:h_1,\dots,h_s\in{k[x_1,\dots,x_n]}\Big\rbrace.$$
\end{define}
It is not hard to show that $\langle{f_1,\dots,f_s}\rangle$ is an ideal. We call it the \emph{ideal generated by} $f_1,\dots,f_s$ and we call $f_1,\dots,f_s$ the \emph{basis} of the ideal.

Studying the relations between certain ideals and varieties forms one of the main areas of study in algebraic geometry. One can even define the algebraic variety $\bold{V}(I)$ defined by the ideal $I\subset{k[x_1,\dots,x_n]}$, where 
$$\bold{V}(I)=\{(a_1,\dots,a_n)\in{k^n}:f(a_1,\dots,a_n)=0,\quad\forall{f\in{I}}\}.$$ The proof that $\bold{V}(I)$ forms an algebraic variety can be found in \cite{geo}. Interestingly, it can also be shown that if $I=\langle{f_1,\dots,f_s}\rangle$, then $\bold{V}(I)=\bold{V}(f_1,\dots,f_s)$. That is to say that \emph{varieties are determined by ideals}. This will have interesting consequences for us, as we will see shortly.



To find a general solution to the implicitization problem introduced in the main text we need to introduce monomial orderings and Groebner bases. 

First, note that we can reconstruct the monomial $x_1^{\alpha_1}\dots{x_n^{\alpha_n}}$ from the $n$-tuple of exponents $(\alpha_1,\dots,\alpha_n)\in\mathbb{Z}_{\geq{0}}^n$. This establishes a one-to-one correspondence between $\mathbb{Z}_{\geq{0}}^n$ and monomials in $k[x_1,\dots,x_n]$. It follows that any ordering $>$ on the space $\mathbb{Z}_{\geq{0}}^n$ will induce an ordering on monomials: if $\alpha>\beta$ according to this ordering, then we will also say that $x^{\alpha}>x^{\beta}$.

Now, we want the induced ordering to be `compatible' with the algebraic structure of the polynomial ring that our monomials live in. This requirement leads us to the following definition.
\begin{define}
A \emph{monomial ordering} on $k[x_1,\dots,x_n]$ is any relation $>$ on $\mathbb{Z}_{\geq{0}}^n$ satisfying:
\begin{enumerate}
\item $>$ is a total ordering on $\mathbb{Z}_{\geq{0}}^n$. That is to say that, for every $\alpha,\beta\in\mathbb{Z}_{\geq{0}}^n$ either $\alpha>\beta$, $\beta>\alpha$ or $\alpha=\beta$.
\item If $\alpha>\beta$ and $\gamma\in\mathbb{Z}_{\geq{0}}^n$, then $\alpha+\gamma>\beta+\gamma$.
\item $>$ is a well ordering on $\mathbb{Z}_{\geq{0}}^n$. This means that every non-empty subset of $\mathbb{Z}_{\geq{0}}^n$ has a smallest element under $>$.
\end{enumerate}
\end{define}

The main monomial ordering we will make use of here is the \emph{lexicographic order}, which we define as follows.

\begin{define}[Lexicographic order]
Let $\alpha=(\alpha_1,\dots,\alpha_n)$ and $\beta=(\beta_1,\dots,\beta_n)\in\mathbb{Z}_{\geq{0}}^n$. We say $\alpha>_{lex}\beta$ if, in the vector difference $\alpha-\beta\in\mathbb{Z}^n$, the leftmost non-zero entry is positive. We will write $x^{\alpha}>_{lex}x^{\beta}$ if $\alpha>_{lex}\beta$.
\end{define}

Once we fix a monomial order, each $f\in{k[x_1,\dots,x_n]}$ has a unique leading term $\tt{LT}\it(f)$ relative to this order. We denote by $\tt{LT}\it(I)$ the set of leading terms of elements of the ideal $I$. We can then define $\langle\tt{LT}\it(I)\rangle$ to be the ideal generated by the elements of $\tt{LT}\it(I)$. Consider a finitely generated ideal $I=\langle{f_1,\dots,f_s}\rangle$, it is interesting to note that $\langle{\tt{LT}}\it(f_1),\dots,\tt{LT}\it(f_s)\rangle$ and $\langle\tt{LT}\it(I)\rangle$ may in general be different ideals. But surprisingly there always exists \cite{geo} a choice of basis $g_1,\dots,g_t\in{I}$ such that $\langle{\tt{LT}}\it(g_1),\dots,\tt{LT}\it(g_t)\rangle=\langle\tt{LT}\it(I)\rangle$. These bases are know as \emph{Groebner bases}.

\begin{define}
Fix a monomial ordering. A finite subset $G=\{g_1,\dots,g_t\}$ of an ideal $I$ is said to be a \emph{Groebner basis} if $$\langle{\tt{LT}}\it(g_1),\dots,\tt{LT}\it(g_t)\rangle=\langle\tt{LT}\it(I)\rangle.$$
\end{define}
 
More informally, a set $G=\{g_1,\dots,g_t\}\subset{I}$ is a Groebner basis for $I$ if and only if the leading term of any element of $I$ is divisible by (at least) one of the $\tt{LT}\it(g_i)$. Groebner bases simplify performing many calculations in algebraic geometry and they have many interesting properties, some of which we will see shortly. There are efficient algorithms for calculating Groebner bases and many software packages that one can use to implement them. \color{black} An example of a Groebner basis was given in the main text. \color{black} Our use of the Groebner basis in that example \color{black} was a special case of a general result, known as the \emph{elimination theorem}, which provides us with a way of using Groebner bases to systematically eliminate certain variables from a system of polynomial equations and thereby solve the implicitization problem. We will state the elimination theorem shortly. First, we require the following definition. \color{black}


\begin{define}
Given $I=\langle{g_1,\dots,g_t}\rangle\subset{k[x_1,\dots,x_n]}$, the \emph{$l^{th}$ elimination ideal} $I_l$ is the ideal of $k[x_1,\dots,x_n]$ defined by 
$$I_l=I\cap{k[x_{l+1},\dots,x_n]}.$$
\end{define}

Thus $I_l$ consists of all consequences of $g_1=\dots=g_t=0$ which eliminate the variables $x_1,\dots,x_l$. Using this language, we see that eliminating $x_1,\dots,x_l$ means finding non-zero polynomials in the $l^{th}$ elimination ideal of $k[x_{l+1},\dots,x_n]$. With the proper ordering, Groebner bases allow us to do this instantly. We can now state the elimination theorem (for a proof, see \cite{geo}).

\begin{thm}[Elimination theorem]
Let $I\subset{k[x_1,\dots,x_n]}$ be an ideal and let $G$ be a Groebner basis for $I$ with respect to the lex order where $x_1>x_2>\dots>x_n$. Then, for every $0\leq{l}\leq{n}$, the set
$$G_l=G\cap{k[x_1,\dots,x_n]}$$
is a Groeber basis of the $l^{th}$ elimination ideal.
\end{thm}

\color{black} So, in our example with the fan depicted in Fig.1(b)---discussed in the main text---$g_3$ and $g_4$ \color{black} form a Groebner basis of the $2^{nd}$ elimination ideal and this is what allowed us to eliminate the variables $q_1$ and $q_2$.

How do we know that we can extend solutions from the $l^{th}$ elimination ideal to the $(l-1)^{th}$? More concretely, in our specific example of the fan, how do we know that the equation $p_{00}p_{01}=p_{10}p_{11}$ defines the entire algebraic variety and not just some part of it? The following result shows us the conditions under which we can extent partial solutions to full ones.

\begin{thm}[Extension theorem]
Let $I\subset{\mathbb{C}[x_1,\dots,x_n]}$ and let $I_1$ be the first elimination ideal of $I$. For each $1\leq i \leq s$, write $f_i$ in the form
$$ f_i=g_i(x_2,\dots, x_n)x_1^{N_i}+\text{terms of lower degree},$$
where $N_i\geq 0$ and $g_i\in{\mathbb{C}[x_1,\dots,x_n]}$ is non-zero. Suppose we had a partial solution $(a_2,\dots,a_n)\in\bold{V}(I_1)$. If $(a_2,\dots,a_n)\notin\bold{V}(g_1,\dots,g_s)$, then there exists $a_1\in\mathbb{C}$ such that $(a_1,\dots,a_n)\in\bold{V}(I)$.
\end{thm} 

\begin{figure}[t]
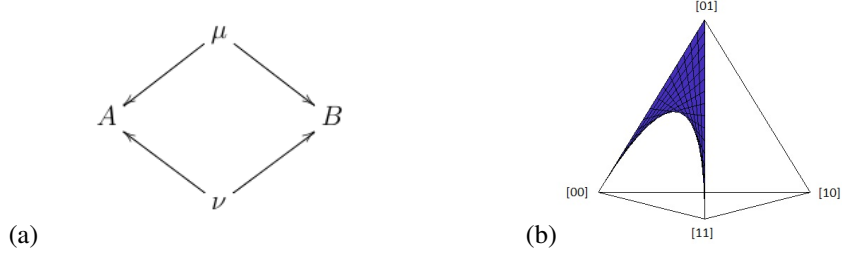

\qquad\qquad\qquad\qquad 
\begin{subfigure}
\centering
(a)\includegraphics[scale=0.5]{DAGst}
\end{subfigure}
\qquad\qquad
\begin{subfigure}
\centering 
(b)\includegraphics[scale=0.35]{Startrek1}
\end{subfigure}
 
\caption{(a) $A=\mu\nu$ and $B=\mu\oplus\nu\oplus\mu\nu$. $\quad$ (b) $\big(p_{01}+2p_{00}\big)^2\geq 4p_{00}$.}
\label{startrek}
\end{figure}

When we work over $(0,1)\subset\mathbb{R}$ we also, in conjunction with the conditions of the above theorem, need to ensure that at every extension step the new solution is real and lies in $(0,1)$. 

We can apply the above theorem to our example to see that, indeed, the equation $p_{00}p_{01}=p_{10}p_{11}$ defines the \color{black} smallest algebraic variety that contains the semi-algebraic set \color{black}
 depicted in Fig.1(b) in the main text.



\section{More examples of deriving tests for feasibility} \label{examples}

Consider the functional causal structure of Fig.~\ref{startrek}(a). The joint distributions that can arise from it are of the form
$$\mathbb{P}(A,B)=q_1q_2[00]+(q_1\overline{q}_2+\overline{q}_1q_2)[01]+\overline{q}_1\overline{q}_2[11].$$ The semi-algebraic set defined by $\mathbb{P}(A,B)$ is shown in Fig.~\ref{startrek}(b). We refer to this variety as a \emph{StarFleet insignia}. The Groebner basis for the ideal 
$$\langle{p_{00}-q_1q_2,p_{01}-q_1\overline{q}_2-\overline{q}_1q_2,p_{11}-\overline{q}_1\overline{q}_2}\rangle,$$
with respect to the lex order $q_1>q_2>p_{00}>p_{01}>p_{11},$ is
$$\begin{aligned}
g_1&=q_1+q_2+p_{00}+2p_{01}-2\\
g_2&=p_{00}+p_{01}+p_{11}-1\\
g_3&=q_2^2+2p_{11}q_2+p_{01}q_2-2q_2-p_{11}-p_{01}+1.
\end{aligned}$$

The equation $g_2=0$ \color{black} defines an equality constraint that restricts the joint probability distribution to the plane $p_{10}=0$ and therefore to \color{black} 
the face of the tetrahedron containing the vertices $[00],[01]$ and $[11].$ In order to extend the partial solution $\{p_{00},p_{01},p_{11}\}$ to a full solution $\{q_1,q_2,p_{00},p_{01},p_{11}\}$ using the extension theorem, we must ensure that all the solutions are real. Now the equation $g_3=0$ allows us to write $q_2$ in terms of the $p_{ij}$'s as follows
$$q_2=\frac{-(p_{01}+2p_{11}-2)\pm\sqrt{(p_{01}+2p_{11}-2)^2+4(p_{11}+p_{01}-1)}}{2}.$$
So in order to ensure that $q_2\in\mathbb{R}$, we must set $(p_{01}+2p_{11}-2)^2+4(p_{11}+p_{01}-1)\geq{0}.$ Using the normalisation condition and rearranging gives us $$(p_{01}+2p_{00})^2\geq{4}p_{00},$$ which defines the semi-algebraic set depicted in Fig.~\ref{startrek}(b). None of the remaining constraints $0 \leq q_i \leq 1$, for $i=1,2,3$ result in non-trivial relations among the $p_{ij}$'s. 

\begin{figure}[t]
\begin{centering}
\begin{subfigure}
(a)\includegraphics[scale=0.58]{DAGnew1}
\end{subfigure}
\begin{subfigure}
~
(b)\includegraphics[scale=0.35]{Thickfan1}  
\end{subfigure}
\begin{subfigure}
~
\includegraphics[scale=0.29]{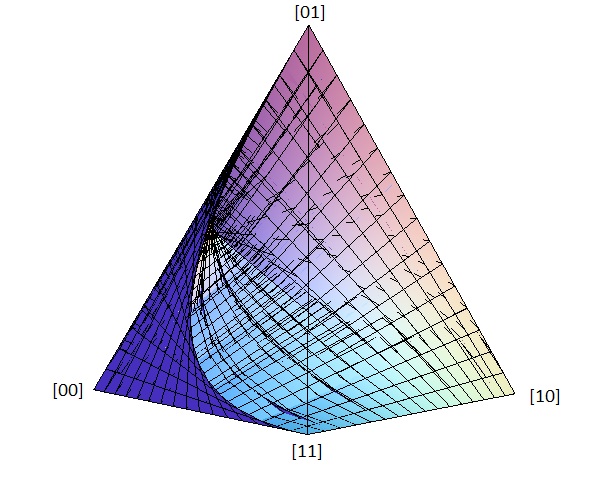}
\end{subfigure}

\caption{(a)  $A=\mu\nu$ and $B=\mu\oplus\nu\oplus\delta$.    (b) $\vert{4}(p_{10}-p_{11})(p_{00}p_{10}-p_{01}p_{11})\vert\leq\big(p_{11}(2p_{01}+2p_{10}+p_{00})-p_{10}(2p_{00}+2p_{11}+p_{01})\big)^2$ }
\label{appendix fig 2}
\end{centering}
\end{figure}

Consider the functional causal structure of Fig.~\ref{appendix fig 2}(a). The joint distributions that can arise from it are of the form
$$ \begin{aligned}
\mathbb{P}(A,B)=(q_1q_2q_3&+\bar{q}_1q_2\bar{q}_3+q_1\bar{q}_2\bar{q}_3)[00]\\
+&(q_1q_2\bar{q}_3+q_1\bar{q}_2q_3+\bar{q}_1q_2q_3)[01] +\bar{q}_1\bar{q}_2q_3[10]+\bar{q}_1\bar{q}_2\bar{q}_3[11].
\end{aligned}$$ 
The semi-algebraic set defined by $\mathbb{P}(A,B)$ is shown from different angles in Fig.~\ref{appendix fig 2}(b). We note that conditioning on the variable $\delta$ being equal to $0$ or $1$ reduces this variety to one of the star trek symbols depicted on the faces. Similarly conditioning on $\nu=1$ (or $\mu=1$) reduces this variety to a fan. 

The Groebner basis for the ideal  
$$\langle{p_{00}-q_1q_2q_3-\bar{q}_1q_2\bar{q}_3-q_1\bar{q}_2\bar{q}_3,
p_{01}-q_1q_2\bar{q}_3-q_1\bar{q}_2q_3-\bar{q}_1q_2q_3,
p_{10}-\bar{q}_1\bar{q}_2q_3,p_{11}-\bar{q}_1\bar{q}_2\bar{q}_3}\rangle,$$
with respect to the usual lex order, is given by  
$$\begin{aligned}
g_1&=q_3p_{10}+q_3p_{11}-p_{10}\\
g_2&=p_{00}+p_{01}+p_{10}+p_{11}-1\\
g_3&=q_2q_1-q_1-q_2-p_{10}-p_{11}+1\\
g_4&=2q_3q_1-q_1-q_2+2q_2q_3-3q_2q_3-3q_3+p_{01}+2p_{10}-p_{11}+1 \\
g_5&=2q_3q_2^2-q_2^2-3q_3q_2+p_{01}q_2+2p_{10}q_2-p_{11}q_2+q_2+q_3-p_{01}-p_{10}\\
g_6&=2p_{10}^2+q_1p_{10}+q_2p_{10}+p_{10}p_{01}+p_{11}p_{10}-2p_{10}-p_{11}^2-q_1p_{11}-q_2p_{11}+p_{01}p_{11}+p_{11}\\
g_7&=p_{10}q_2^2-p_{11}q_2^2+2p_{10}^2q_2-p_{11}^2q_2+p_{01}p_{10}q_2-2p_{10}q_2\\
&\qquad\qquad+p_{01}p_{11}q_2+p_{10}p_{11}q_2+p_{11}q_2-p_{10}^2-p_{01}p_{10}+p_{10}-p_{01}p_{11}-p_{10}p_{11}.
\end{aligned}$$
The equation $g_2=0$ is just the usual normalisation condition \color{black} restricting the joint probability distribution to the tetrahedron.\color{black} In order to use the extension theorem to extend a partial solution $\{p_{00},p_{01},p_{10},p_{11}\}$ to a full solution $\{q_1,q_2,q_3,p_{00},p_{01},p_{10},p_{11}\}$, we must ensure that each solution is real. The only situations in which we need to impose this is in the case of $q_2$. The equation $g_7=0$ is a quadratic in $q_2$ and in order for its solutions to be real, we must stipulate that
$$4(p_{10}-p_{11})(p_{00}p_{10}-p_{01}p_{11})\leq\big(p_{11}(2p_{01}+2p_{10}+p_{00})-p_{10}(2p_{00}+2p_{11}+p_{01})\big)^2.$$
Using $g_1=0$ to write $q_3$ in terms of $p_{10}$ and $p_{11}$ and substituting this into $g_5=0$ gives us another quadratic in $q_2$. For the solutions of this quadratic to be real we must enforce that
$$4(p_{10}-p_{11})(p_{00}p_{10}-p_{01}p_{11})\geq{-}\big(p_{11}(2p_{01}+2p_{10}+p_{00})-p_{10}(2p_{00}+2p_{11}+p_{01})\big)^2.$$
Combining these two inequalities we get 
$$\vert{4}(p_{10}-p_{11})(p_{00}p_{10}-p_{01}p_{11})\vert\leq\big(p_{11}(2p_{01}+2p_{10}+p_{00})-p_{10}(2p_{00}+2p_{11}+p_{01})\big)^2,$$
where $\vert.\vert$ denotes the absolute value. None of the remaining constraints $0 \leq q_i \leq 1$, for $i=1,2,3$ result in non-trivial relations among the $p_{ij}$'s.  

Examining this inequality more closely, we see that setting $p_{10}=0$ reduces it to the inequality defining the StarFleet insignia, $4p_{01}\leq(2p_{01}+p_{00})^2$, on the face spanned by $\{[00],[01],[11]\}$, as it should (this is visible in Fig.~\ref{appendix fig 2}(b)). Similarly, for $p_{11}=0$ we get the StarFleet insignia on the face $\{[00], [01],[10]\}$ (also visible in Fig.~\ref{appendix fig 2}(b)). The appearance of the term $p_{00}p_{10}-p_{01}p_{11}$ is also noteworthy. Recall that the equation $p_{00}p_{10}=p_{01}p_{11}$ defines the fan depicted in Fig.2(b) in the main text, so the above inequality quantitatively bounds the deviation from the surface of this fan by an amount proportional to the two star trek symbols discussed above. This is intuitively what we would expect from looking at the semi-algebraic set depicted in Fig.~\ref{appendix fig 2}(b). 

These examples cover all the different situations one may encounter while using algebraic geometry techniques to derive tests for feasibility of the causal models we are considering in this work. The remaining tests are derived in an analogous fashion.

\section{Sufficiency of $n$-latent-bit models}  \label{sufficientcy} 

We here present the proof of Theorem~\ref{thm:binarylatents}.

The example presented in section~\ref{binary-latents} suggests a general procedure for replacing an $m$-valued latent variable with some finite number of binary latent variables. Replace the $m$-valued variable with a number of {\em substitute} variables---the analogues of $\gamma$ and $\eta$ above, but which now can take an arbitrary number of values---such that any distribution over the $m$-valued variable can be simulated using a $k$-latent-bit causal model---the analogue of the causal model containing $\mu$ and $\nu$ above---underlying the substitute variables. By eliminating the intermediary variables, the dependence of the observed variables on the $m$-valued latent variable is replaced with a dependence on $k$ binary latent variables. 



We now describe a procedure for replacing an $m$-valued variable, for any $m$, by two variables $\gamma$ and $\eta$ in such a way that any distribution over the $m$-valued variable is obtained  by some $k$-latent-bit causal model underlying $\gamma$ and $\eta$.  

Recall that for a 3-valued variable, we can take $\gamma$ and $\eta$ to be bits and use the fiducial model from class $(2,1,c)_{\rm{Id}}$, whose distribution is the convex combination of an edge of the tetrahedron and a vertex not contained in that edge.  Similarly, for a 4-valued variable, we can take $\gamma$ and $\eta$ to be bits and use the fiducial model from class $(3,2,g)_{\rm{Id}}$, whose distribution is the convex combination of a face and vertex not contained in that face. 

For a 5-valued variable, we can take $\gamma$ to be a trit and $\eta$ to be a bit.
For any causal model underlying $\gamma$ and $\eta$, the semi-algebraic set generated by this model is now a subset of a simplex with six vertices, $[\gamma \eta] \in \{ [00],[01],[10],[11],[20],[21]\}$
We now construct a causal model underlying $\gamma$ and $\eta$ by combining two simpler models, using the procedure described in section 4 in the main text: the first model is one whose semi-algebraic set is the tetrahedron (considered as the subset of the six-simplex having 
$[\gamma \eta] \in \{[00], [01],[10],[11]\}$)
 and the second is one whose semi-algebraic set is a vertex of the six-simplex not contained in the tetrahedron. 
A binary switch variable toggles between these two simpler models.  Given the geometry, the semi-algebraic set defined by the model is clearly the convex combination of the tetrahedron and the vertex outside the tetrahedron.  
In particular, we can take the first model to be the fiducial model from class $(3,3)_{\rm{Id}}$ (where $\gamma$ is replaced by a trit but its dependence on its causal parents is unchanged) and the second model to be a deterministic model that sets $\gamma=2$ and $\eta=0$.   Denoting the switch variable by $\rho$, and the other latent bits by $\mu$, $\nu$, $\delta$ (as in the row containing class $(3,3)_{\rm{Id}}$), we obtain the following functional dependences by the switch-variable construction:
 $\gamma =\rho(\mu \nu\oplus_2 1)\oplus_3 2(\rho\oplus_2 1)$ and $\eta =\rho(\mu\nu\delta\oplus_2\nu).$   
One easily verified that if $\rho=1$, one recovers the fiducial model of class $(3,3)_{\rm{Id}}$ and hence the tetrahedron spanned by $[\gamma \eta] \in \{ [00],[01],[10],[11]\}$, while if $\rho=0$, one obtains the point $[\gamma \eta] = [20]$.

By increasing the number of values that $\gamma$ and $\eta$ can take, one can ensure that the number of vertices in the space of distributions over $\gamma$ and $\eta$ is at least $m$, such that one can simulate an $m$-valued latent variable by finding a causal model underlying $\gamma$ and $\eta$ whose semi-algebraic set is an $m$-simplex.  
To construct such a model, we apply the switch-variable construction to a pair of simpler models, one of which has an semi-algebraic set corresponding to an $(m-1)$-simplex, and the other of which is a deterministic model corresponding to a vertex outside of this $(m-1)$-simplex.  In this way, we can recursively build up a causal model involving only binary latent variables whose semi-algebraic set is an $m$-simplex for any $m$. 



 

\end{appendix}

\end{document}